\pdfoutput=1

\documentclass[11pt]{article}

\usepackage[]{EMNLP2023}

\usepackage{times}
\usepackage{latexsym}

\usepackage[T1]{fontenc}

\usepackage[utf8]{inputenc}

\usepackage{microtype}

\usepackage{inconsolata}

\usepackage{paralist}
\usepackage{graphicx}
\usepackage{float}
\usepackage{subcaption}
\usepackage{booktabs}
\usepackage{adjustbox}
\usepackage{multirow}

\usepackage{xcolor}
\usepackage{multirow}
\usepackage{adjustbox}        
\usepackage{colortbl}         
\usepackage{amssymb}       
\usepackage{amsmath}

\usepackage{xcolor, soul}
\sethlcolor{cyan}

\usepackage{cleveref}
\crefname{section}{\S}{\S\S}
\crefname{table}{Table}{}
\crefname{figure}{Figure}{}
\crefname{appendix}{Appendix}{}
\crefformat{section}{\S#2#1#3}

\newcommand{\hlc}[2][yellow]{{%
    \colorlet{foo}{#1}%
    \sethlcolor{foo}\hl{#2}}%
}

\title{The Validity of Evaluation Results: Assessing Concurrence Across Compositionality Benchmarks
}

 \author{Kaiser Sun \quad Adina Williams \quad Dieuwke Hupkes \\
Meta AI\\
\texttt{hsun74@cs.jhu.edu} \\ \texttt{\{adinawilliams, dieuwkehupkes\}@meta.com}\\
}

\begin{document}
\maketitle

\begin{abstract}
NLP models have progressed drastically in recent years, according to numerous datasets proposed to evaluate performance.
Questions remain, however, about how particular dataset design choices may impact the conclusions we draw about model capabilities.
In this work, we investigate this question in the domain of compositional generalization. 
We examine the performance of six modeling approaches across 4 datasets, split according to 8 compositional splitting strategies, ranking models by 18 compositional generalization splits in total.
Our results show that:
i) the datasets, although all designed to evaluate compositional generalization, rank modeling approaches differently;
ii) datasets generated by humans align better with each other than they with synthetic datasets, or than synthetic datasets among themselves;
iii) generally, whether datasets are sampled from the same source is more predictive of the resulting model ranking than whether they maintain the same interpretation of compositionality; and
iv) which lexical items are used in the data can strongly impact conclusions.
Overall, our results demonstrate that much work remains to be done when it comes to assessing whether popular evaluation datasets measure what they intend to measure, and suggests that elucidating more rigorous standards for establishing the validity of evaluation sets could benefit the field.\footnote{Code to reproduce the experiments can be found at \url{https://github.com/facebookresearch/CompositionalityValidity}.}
\end{abstract}

\section{Introduction}

\begin{figure}[t]
\centering
\includegraphics[width=\the\columnwidth]{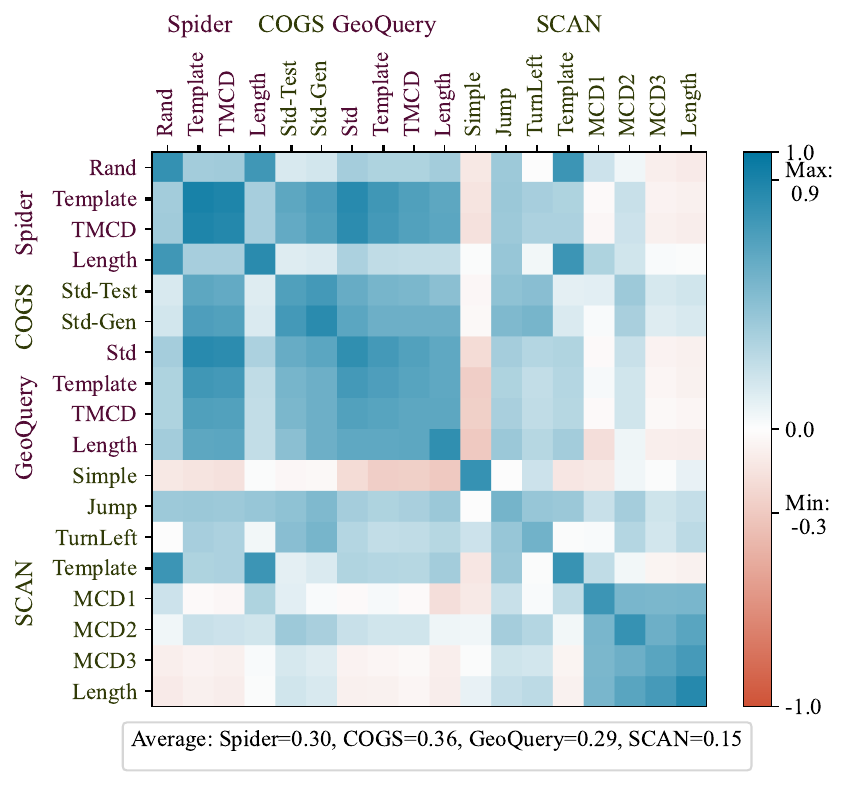}
    \caption{Pairwise concurrence values averaged across models for each dataset--split pair. Values closer to 1.0 (blue) denote a more similar ranking of models according to their performance on the dataset and split. The dataset and split font color indicate whether the data was generated by humans (purple) or synthetically using rules (green).}
\label{fig:default_conf_matrix}
\end{figure}

\begin{figure}
\centering
\includegraphics[width=0.8\columnwidth]{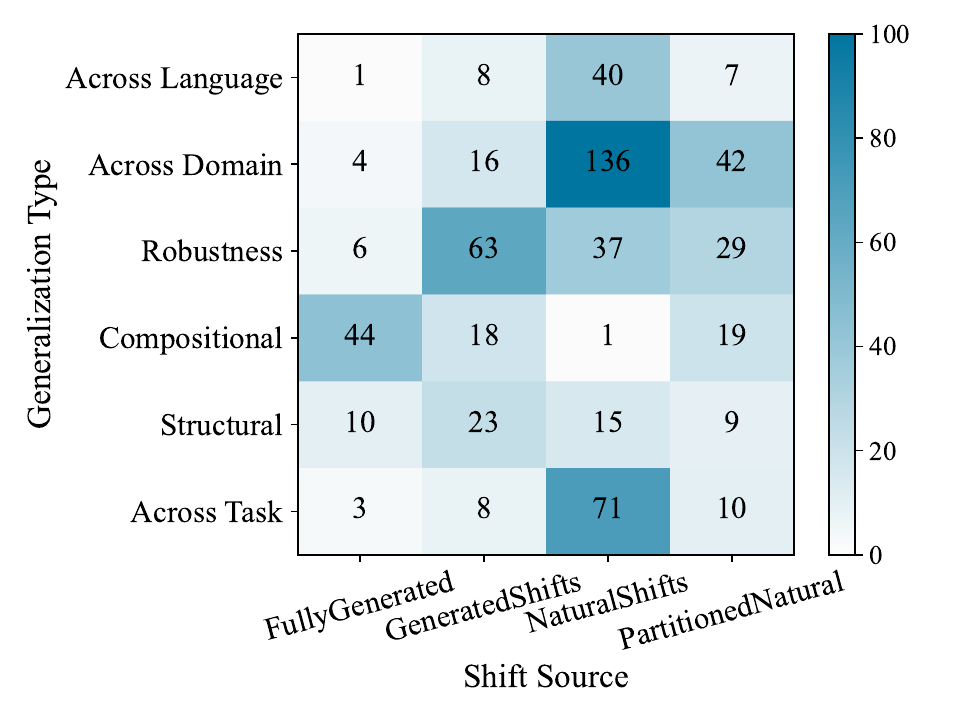}
    \caption{Generalization studies published in the ACL anthology (2015-2022), across different \emph{shift sources}.}\label{fig:generalisation_vs_source}
\end{figure}

Over the past few years, NLP has made astonishing progress on almost all language-related tasks proposed by the community. %
Concurrently, %
a plethora of benchmark datasets has emerged for evaluating the skills of NLP models and exposing their strengths and weaknesses (\citealt{chowdhery-etal-2022-scaling}, \textit{inter alia}).
These datasets focus on a variety of different aspects of model capabilities, that are increasingly not mutually exclusive: oftentimes, multiple benchmarks are available that target the same capability or skill, using (slightly) different metrics, design choices, and/or conceptual approaches. 
For instance, \citet{Hupkes2023} report that many recent studies on generalization  used  different \emph{shift sources} to study the same types of \emph{generalization} (see \cref{fig:generalisation_vs_source}).\footnote{Plot generated using the visualisation tool on \url{https://genbench.org/visualisations}.} 

However, somewhat surprisingly, despite a wealth of work in the domain of evaluation and generalization, there is very little research that assesses whether multiple datasets designed to measure the same ability also yield the same conclusions.
This makes it difficult for practitioners to conduct informed evaluation dataset selection and, perhaps even more concerning, impedes our understanding of how well different datasets measure what they intend to measure.
While establishing \emph{construct validity} and \emph{construct reliability} -- for instance through comparing the results of tests with other tests that intend to measure the same thing -- is common practice in the social sciences \citep{westen2003quantifying, jacobs-wallach-2021-measurement}, it is not the standard in the field of NLP.

In this work, we argue that establishing such standards is much needed in our field, and we present a detailed set of experiments that assesses construct validity in the domain of \emph{compositional generalization}. 
Following \citet{liu2021question}, we use \emph{concurrence} to measure the extent to which 8 different \emph{compositional splitting strategies} for 4 different datasets -- SCAN, GeoQuery, COGS, and Spider -- %
provide similar rankings for 6 different modeling approaches -- BART, T5, Transformer, uni- and biLSTMS, and Neural-BTG.
We find that, in general, the conclusions drawn from one dataset split typically do not align with the results from another dataset split.
In a range of experiments, we explore if that could be attributed to whether the underlying data are synthetic or human-generated, to the compositional splitting strategy is used to create the data (a.k.a.\ what interpretation of compositionality), or to uncontrolled exposure to lexical items that also occurred during pretraining.

We find that concurrence values are generally low: only 10 out of 153 pairs of dataset splits have a concurrence value that surpasses the threshold for high concurrence.
Furthermore, results from human-authored datasets concur much more than results from synthetic datasets.
On the contrary, dataset splits that share the same interpretation of compositionality -- as defined by their splitting strategy -- hardly concur with each other:  the underlying data plays a more important role in model rankings.
Lastly, aligned with the findings of \citet{kim2022uncontrolled}, we find that carefully controlling the lexical items in a compositional split has a positive impact on concurrence.
Overall, our results suggest that much work remains to be done to evaluate compositional generalization, and more generally that having more rigorous standards for establishing the validity of evaluation sets should be prioritized in the future.

\section{Related Work}
In this section, we provide an overview of datasets commonly used for assessing compositional generalization, and we discuss previous attempts to compare performance across benchmarks.

\paragraph{Datasets for Compositional Generalization}
Since the introduction of \emph{SCAN} in 2018 \citep{lake2018generalization}, many datasets have been proposed to assess compositional generalization in neural networks.
Several of them were direct follow-ups to SCAN that aimed to extend the original dataset or mitigate various issues perceived with it.
For instance, \citet{bastings-etal-2018-jump} introduced NACS, a `reversed' version of SCAN; \citet{loula-etal-2018-rearranging} introduced new splits using the original dataset; \citet{ruis-etal-2020-gscan} introduced a multimodal, grounded version of the benchmark; and \citet{patel-etal-2022-revisiting} increased the number of primitives.
Recently, \citet{valvoda-etal-2022-benchmarking} proposed a transducer-based procedure for generating myriad synthetic datasets similar to SCAN to investigate which formal properties impact the results.
Other artificially generated datasets available to evaluate compositionality are PCFG SET \citep{hupkes2020compositionality}, %
COGS \citep{kim-linzen-2020-cogs}, %
and the dataset proposed by \citet{oren-etal-2021-finding}.

Datasets that use more natural (but often still templated) data are typically situated in the domain of machine translation -- such as \citet{li-etal-2021-compositional}, \citet{dankers-etal-2022-paradox} and \citet{raunak2019compositionality} -- or semantic parsing -- e.g.\ \citet{finegan2018improving,keysers2019measuring,shaw-etal-2021-compositional,cui-etal-2022-compositional}.
Finally, \citet{thrush-etal-2022-winoground} introduce Winoground, aimed to assess compositionality in text-to-image models.
In our work, we focus on datasets that target compositionality in the domain of semantic parsing, with the addition of SCAN for its sheer popularity.

\paragraph{Performance across benchmarks}
Several recent works across NLP have been interested in the extent to which strong performance on one task, setting, or dataset transfers to strong performance on another.
Typically, such experiments are motivated by transfer learning, rather than establishing the validity of evaluation results.
For instance, \citet{vu-etal-2020-exploring}, \citet{ye-etal-2021-crossfit}, \citet{luo-etal-2022-cogtaskonomy}, \citet{padmakumar-etal-2022-exploring}, and \citet{weber-etal-2021-language} all investigate to what extent performance transfers across tasks.
More closely related to our study, is the work presented by \citet{liu2021question}, who quantify the measurement of benchmark agreement on model rankings and compare it in question answering.
In our work, we adopt their definition of comparability across datasets.

In the context of compositional generalization, the work most closely related to ours is the study presented by \citet{chaabouni-etal-2021-transformers}, in which they investigate whether the performance improvements on the synthetic dataset SCAN transfer to the naturalistic setting.
We largely confirm their results, but consider compositionality benchmarks more broadly, not only considering the synthetic v.s\ natural dimension, but also interpretations of compositionality and lexical items exposed during pretraining.

\section{Methodology}\label{sec:Methods}
We compare how the conclusions drawn from 18 different compositional generalization splits -- defined over 4 different datasets with 8 compositional splitting strategies -- compare across 6 modeling approaches.
In this section, we describe the datasets and modeling approaches we consider and provide details on training and hyperparameter selection.

\subsection{Models}\label{subsec:models}
For our experiments, we consider both pretrained and train-from-scratch approaches that have previously been considered in the context of compositional generalization.

\paragraph{BART \& T5}
We use the pretrained seq2seq models BART \cite{lewis-etal-2020-bart} and T5 \cite{raffel2020exploring} to enable easy comparison with prior work. 
In the case of BART, order-based noising strategies are used, which may encourage the model to learn to better represent linguistic structure. 

\paragraph{LSTM \& Transformer} 
To ensure coverage of models without pre-trained knowledge, we use a uni-directional LSTM \cite{hochreiter1997long}, a bi-directional LSTM, and a vanilla transformer \cite{vaswani2017attention}.

\paragraph{Neural-BTG}
We include one modeling approach specifically designed to address compositionality: Neural-BTG \cite{wang-etal-2022-hierarchical-phrase}, %
composed of a discriminative parser based on a bracketing transduction grammar (BTG; \citealp{wu-1997-stochastic}) %
and a neural seq2seq model. %

\subsection{Data}\label{subsec:data}
We consider four different datasets designed to test compositional generalization.
We focus on datasets for semantic parsing and include SCAN as the most commonly used dataset for compositionality overall.
Three of these datasets contain different curated \emph{splits} that target different interpretations of compositionality.
Two of the datasets (SCAN and COGS) are synthetic datasets that are generated with rules, while the other two (Spider and GeoQuery) are natural datasets, authored by humans.
Examples for all datasets and descriptions of all curated splits can be found in \cref{app:sec:data}.

\begin{table}[t]
\scriptsize
\centering
\begin{tabular}{p{3em}p{4em}p{0.28\textwidth}}
\toprule
\multirow{2}{*}{\textbf{COGS}} & \texttt{Input:} & Mila liked that the cake was offered to Emma . \\
& \texttt{Output:} & \texttt{* cake ( x \_ 4 ) ; like . agent ( x \_ 1 , Mila ) AND like . ccomp ( x \_ 1 , x \_ 6 ) AND offer . theme ( x \_ 6 , x \_ 4 ) AND offer . recipient ( x \_ 6 , Emma )}
\\ \midrule
\multirow{2}{*}{\textbf{SCAN}}  &  \texttt{Input:} & turn left after jump twice \\
 & \texttt{Output:} & \texttt{I\_JUMP I\_JUMP I\_TURN\_LEFT}
\\ \midrule
\multirow{2}{*}{\textbf{GeoQuery }}  &  \texttt{Input:} & how much population does m0 have \\
&  \texttt{Output:} & \texttt{answer ( intersection ( river , loc\_2 ( m0 ) ) )} \\ \midrule
\multirow{2}{*}{\textbf{Spider}}  &  \texttt{Input:} & flight\_1: what is the average distance and price for all flights from la?  \\
&  \texttt{Output:} & \texttt{select avg(distance) , avg(price) from flight where origin = "los angeles"} \\
\bottomrule
\end{tabular}
\caption{Examples of instances in each dataset used in our experiments.}\label{tab:mainexamples}

\end{table}

\paragraph{SCAN} Consisting of a set of commands and the corresponding action sequences, SCAN \cite{lake2018generalization} is one of the most popular synthetic datasets to study compositional generalization.
We include the \textit{simple}, \textit{length}, \textit{add primitive}, \textit{template} splits from \citet{lake2018generalization}.
In addition to original SCAN splits, we also use the maximum compound divergence (MCD) splits of SCAN proposed by \citet{keysers2020measuring}.

\paragraph{COGS} 
\citet{kim-linzen-2020-cogs} introduced COGS, a synthetic semantic parsing dataset generated by a rule-based approach, which covers a larger variety of grammar rules than SCAN does.
The inputs in COGS are English sentences, generated by a probabilistic context-free grammar. 
The corresponding output, which is the semantic interpretation of the input, is annotated with the logical formalism of \citet{reddy-etal-2017-universal}.
COGS includes a randomly sampled test set and an out-of-distribution compositional generalization set.

\paragraph{GeoQuery} GeoQuery \cite{tang2001using, zelle1996learning} is a text-to-QL dataset containing naturalistic examples.
We use the four compositional generalization splits defined on this dataset by \citet{shaw-etal-2021-compositional}: 
\textit{random/standard}, \textit{length}, \textit{template}, and \textit{Target Maximum Compound Divergence (TMCD)}.

\paragraph{Spider} Spider \cite{yu-etal-2018-Spider} is originally designed for cross-domain semantic parsing. %
We use the compositional generalization splits for Spider defined by \citet{shaw-etal-2021-compositional}, which match their splits for GeoQuery: 
\textit{random/standard}, \textit{length}, \textit{template}, and \textit{TMCD}.

\subsection{Training Setup}
We train/fine-tune the models on the train partition of each dataset described above and evaluate them on the corresponding test set.
For T5 on GeoQuery and Spider as well as LSTM and Transformers on COGS, we use the hyperparameters provided in \citet{shaw-etal-2021-compositional} and \citet{kim-linzen-2020-cogs}, respectively. 
We followed \citet{orhan2021compositional} to train T5 and \citet{yao-koller-2022-structural} to train BART on COGS. 
For the remaining model-dataset combinations, we perform a hyperparameter search for each dataset, with 10\% of instances randomly chosen to be used for tuning. 
Details can be found in \cref{app:hp_tuning}. 
We use three different random seeds for each training run and use five random seeds for each training run of LSTM, to compensate for LSTM's higher variation in performance across seeds.
For models with existing evaluations on a dataset, we compare to these previous measures of performance to ensure that our replication results align with previously reported numbers \cite{keysers2020measuring, kim-linzen-2020-cogs, orhan2021compositional, shaw-etal-2021-compositional, yao-koller-2022-structural, sun-etal-2023-resp}.

\subsection{Evaluation}
\label{sec:expsetup:eval}
For most datasets, we use exact match (EM) accuracy.
EM is a binary metric that only counts an output as correct if it matches the target output exactly, and is most frequently used for the datasets we consider.
During initial experiments, we found that, in many cases, EM accuracy may be too strict for our purposes.
In some cases, models' tokenizers may prefer slightly different spacing -- a phenomenon also reported by \citet{sun2022tokenization} -- in others, models lack specific tokens in their vocabulary.
Neither of these things is indicative of a model's compositional generalization capability, and we therefore choose to normalize model outputs before applying EM accuracy.
In \cref{app:sec:EM}, we include examples of such cases, and we report the differences between EM scores with and without our normalization step.
For Spider, the original dataset also uses a more lenient EM implementation.
For consistency reasons, we use the same implementation across all datasets, but we report Spider EM scores in \cref{app:sec:spider} to compare with previous work. 

\subsection{Measuring Concurrence}
To measure how similarly different dataset splits rank different modeling approaches, we use the concept of \emph{concurrence} introduced by \citet{liu2021question}.
The concurrence between two dataset splits is defined as the correlation between the performances of different modeling approaches for those splits.
More specifically, the concurrence $\text{CONCUR}(D_1, D_2; \mathcal{A}, \text{Eval})$ between two dataset splits $D_1$ and $D_2$, given a set of modeling approaches $\mathcal{A}$ and evaluation function $\text{Eval}$, is defined as:%
\begin{equation*}
    \text{CONCUR}(D_1, D_2; \mathcal{A}, \text{Eval})=\text{CORR}(P_1, P_2),
\end{equation*}
where $\text{CORR}$ is some correlation function and $P_i$ is the variable that holds the scores of Eval($a, D_i$) for all $a\in\mathcal{A}$.
For \text{CORR}, \citet{liu2021question} considered both Pearson ($r$) and Kendall rank ($\tau$). 
Because we are interested in how benchmarks rank model performance, we report the concurrence values under Kendall's $\tau$ unless specified otherwise.
We refer to the concurrence between the dataset split and itself as \emph{self-concurrence}, the value of which is purely affected by seed variation across training runs. 
We see self-concurrence, which would be $1.0$ if there is no variation across seeds, as an upper bound for the concurrence values across dataset splits.

\section{Results}

We now present our results, starting with a discussion of the performance of models on the datasets (\cref{sec:overall_perf}) and the concurrence scores between the performances (\cref{sec:predictivity}), we then proceed to look at the relationship between synthetic and natural compositionality datasets (\cref{sec:synvnat}), and how this interacts with the choice of definition of compositionality and underlying dataset (\cref{sec:comp_def}). We finish our results section with a short investigation into the impact of the choice of lexical items in data (\cref{sec:lex_exposure}).

\begin{table*}[ht]
\centering
\scriptsize
\begin{adjustbox}{max width=\textwidth}
\begin{tabular}{clrlrlrlrlrlrlr}
\toprule
\textbf{Dataset} &
  \textbf{Split} &
  \multicolumn{2}{c}{\textbf{LSTM Uni}} &
  \multicolumn{2}{c}{\textbf{LSTM Bi}} &
  \multicolumn{2}{c}{\textbf{Transformer}} &
  \multicolumn{2}{c}{\textbf{T5}} &
  \multicolumn{2}{c}{\textbf{BART}} &
  \multicolumn{2}{c}{\textbf{BTG}} &
  \multicolumn{1}{c}{\textbf{Avg}} \\ \midrule
\multirow{2}{*}{COGS}     & \textit{Std-Test} & 99.3 & ±.0  & 99.1 & ±.01 & 99.5  & ±.0  & 99.7 & ±.0  & 99.7 & ±.0  & 68.8 & ±.01 & 94.3 \\
                          & \textit{Std-Gen}  & 21.3 & ±.05 & 14.8 & ±.08 & 56.1  & ±.06 & 82.9 & ±.0  & 78.6 & ±.0  & 2.8  & ±.01 & 42.8 \\ \midrule
\multirow{8}{*}{SCAN}     & \textit{Simple}   & 99.9 & ±.0  & 99.9 & ±.0  & 100.0 & ±.0  & 94.9 & ±.01 & 99.1 & ±.01 & 12.3 & ±.01 & 84.4 \\
                          & \textit{Jump}     & 0.4  & ±.01 & 0.0  & ±.0  & 0.1   & ±.0  & 95.0 & ±.01 & 0.4  & ±.01 & 0.0  & ±.0  & 16.0 \\
                          & \textit{TurnLeft} & 61.1 & ±.13 & 34.1 & ±.06 & 64.8  & ±.11 & 70.3 & ±.12 & 63.1 & ±.19 & 8.9  & ±.01 & 50.4 \\
                          & \textit{Template} & 0.2  & ±.0  & 0.3  & ±.01 & 1.1   & ±.0  & 34.3 & ±.03 & 0.0  & ±.0  & 0.9  & ±.01 & 6.1  \\
                          & \textit{MCD1}     & 5.9  & ±.06 & 12.2 & ±.07 & 1.1   & ±.0  & 24.6 & ±.01 & 0.4  & ±.01 & 1.8  & ±.01 & 7.7  \\
                          & \textit{MCD2}     & 6.7  & ±.03 & 5.8  & ±.03 & 1.2   & ±.0  & 34.1 & ±.01 & 1.6  & ±.0  & 0.5  & ±.0  & 8.3  \\
                          & \textit{MCD3}     & 8.7  & ±.04 & 7.8  & ±.02 & 0.7   & ±.0  & 11.1 & ±.01 & 1.2  & ±.01 & 0.8  & ±.01 & 5.0  \\
                          & \textit{Length}   & 15.3 & ±.04 & 11.8 & ±.01 & 0.0   & ±.0  & 14.1 & ±.01 & 0.7  & ±.01 & 0.0  & ±.0  & 7.0  \\ \midrule
\multirow{4}{*}{GeoQuery} & \textit{Std}      & 74.0 & ±.06 & 78.9 & ±.04 & 82.3  & ±.02 & 92.5 & ±.01 & 89.2 & ±.01 & 79.0 & ±.01 & 82.6 \\
                          & \textit{Template} & 46.5 & ±.06 & 55.9 & ±.07 & 56.7  & ±.04 & 91.0 & ±.0  & 77.1 & ±.06 & 53.5 & ±.06 & 63.5 \\
                          & \textit{TMCD}     & 35.8 & ±.02 & 37.1 & ±.02 & 37.9  & ±.01 & 54.1 & ±.0  & 48.2 & ±.0  & 36.9 & ±.0  & 41.7 \\
                          & \textit{Length}   & 18.5 & ±.03 & 16.2 & ±.02 & 22.0  & ±.01 & 41.1 & ±.01 & 36.1 & ±.01 & 20.7 & ±.02 & 25.8 \\ \midrule
\multirow{4}{*}{Spider}   & \textit{Rand}     & 33.4 & ±.02 & 36.9 & ±.01 & 42.5  & ±.01 & 68.0 & ±.0  & 32.7 & ±.01 & 40.1 & ±.01 & 42.3 \\
                          & \textit{Template} & 1.0  & ±.0  & 2.2  & ±.01 & 4.6   & ±.0  & 39.6 & ±.01 & 21.6 & ±.01 & 1.9  & ±.0  & 11.8 \\
                          & \textit{TMCD}     & 4.6  & ±.01 & 6.0  & ±.01 & 7.5   & ±.01 & 47.2 & ±.01 & 31.2 & ±.03 & 5.5  & ±.0  & 17.0 \\
                          & \textit{Length}   & 12.7 & ±.01 & 14.0 & ±.01 & 17.5  & ±.01 & 35.4 & ±.01 & 7.4  & ±.0  & 14.0 & ±.01 & 16.8
 \\
                        \bottomrule
\end{tabular}
\end{adjustbox}
    \caption{Model exact-match accuracy on datasets averaged across random seeds, with standard deviation.}%
\label{tab:perfStd}
\end{table*}

\subsection{Overall Performance}
\label{sec:overall_perf}
In \cref{tab:perfStd}, we show the performance of all models on all dataset splits under consideration, as well as the average performance per dataset split (last column).
Our scores are generally close to the scores reported in previous work, for the (dataset split, architecture) combinations for which previous results exist \citep{sun-etal-2023-resp}, with the exception of the results for Spider, for which we use a different metric.
All models perform reasonably well on the random splits of each datasets (first row for each dataset in \cref{tab:perfStd}), but most struggle with various generalization splits. 
While some splits are difficult across the board, other difficulties appear more model-dependent.
For instance, while all models are weak on the \textit{length} and \textit{MCD} splits of SCAN and \textit{length} split of Spider, COGS is difficult for some models (e.g., BTG) but much less for others (e.g., T5).
Similarly, some models perform well on one of the datasets or one of the splits, but perform poorly on the others. 
BART, for instance, maintains high performance on GeoQuery and COGS, but performs even worse than non-pretrained models on some splits of SCAN, while BTG performs well on GeoQuery but fails on many splits of SCAN.
T5 has high performance on most datasets, but is outperformed by the unidirectional LSTM on the \textit{length} split of SCAN.
SCAN, in particular, appears to be challenging for all models, with the \textit{TurnLeft} split being the only exception.%
\footnote{While architectures exist that obtain high scores on SCAN, such as the ones introduced by \citet{shaw-etal-2021-compositional} and \citet{kim2021sequencetosequence}, they are too narrowly scoped for our current study and we thus do not consider them.}

\subsection{Overall Concurrence}
\label{sec:predictivity}
\begin{table}[t]
\centering
\scriptsize
\begin{tabular}{llllc}
\toprule
\textbf{Dataset A} & \textbf{Dataset B} & \textbf{Split A} & \textbf{Split B} & \textbf{Concur} \\ \midrule
Spider   & Spider   & \textit{Template} & \textit{TMCD}     & 0.88 \\
GeoQuery & Spider   & \textit{Std}      & \textit{Template} & 0.84 \\
GeoQuery & Spider   & \textit{Std}      & \textit{TMCD}     & 0.83 \\
SCAN     & Spider   & \textit{Template} & \textit{Rand}     & 0.76 \\
SCAN     & Spider   & \textit{Template} & \textit{Length}   & 0.76 \\
Spider   & Spider   & \textit{Rand}     & \textit{Length}   & 0.75 \\
GeoQuery & Spider   & \textit{Template} & \textit{Template} & 0.74 \\
GeoQuery & Spider   & \textit{Template} & \textit{TMCD}     & 0.73 \\
GeoQuery & GeoQuery & \textit{Std}      & \textit{Template} & 0.73 \\
SCAN     & SCAN     & \textit{Length}   & \textit{MCD3}     & 0.72 \\
\bottomrule
\end{tabular}
\caption{High concurrence values ($\geq 0.7$) among all pairs of dataset splits, excluding self-concurrence.}
\label{tab:concurrence}
\end{table}

It is not difficult to tell from Table~\ref{tab:perfStd} that the performance of a model on one dataset is not predictive of its performance on the others.  
To quantitatively substantiate this observation, we compute the concurrences between the different dataset splits, which we visualize in \cref{fig:default_conf_matrix}.
On average, the concurrence between dataset splits is low: a mere $0.22$, far below the average self-concurrence of $0.76$ that (model, split) combinations have across different seeds.
Interestingly, even these average self-concurrence values are lower than the $0.8$ that \citet{liu2021question} used as a threshold for ``high'' concurrence, indicating that performance on the same compositional dataset is not very stable across runs.\footnote{This finding is in line with results reported by \citet{liska2018memorize}, who find a range of different generalization performances on a simple but highly compositional look-up table task.}
Consequently, we lower the threshold to $0.7$ here, which is approximately 90\% of the average self-concurrence.
Of the 153 pairs of dataset split we compare in this experiment, only 10 pairs surpass this threshold.
Somewhat surprisingly, perhaps, many of the highest values (reported in \cref{tab:concurrence}), are concurrences between i.i.d.\ splits and compositional splits.

Considering the concurrence of each dataset with all other datasets (excluding self-concurrence, values are reported below \cref{fig:default_conf_matrix}), we can see that performance COGS, with an average $\tau$ of $0.36$ is most predictive of performance on other datasets.
Furthermore, the three semantic parsing datasets have much higher average concurrence than SCAN, suggesting that compositionality on one task may not be predictive of compositionality on another.

\subsection{Synthetic vs natural data}\label{sec:synvnat}
\begin{figure}[t]
\centering
\includegraphics[width=0.9\columnwidth]{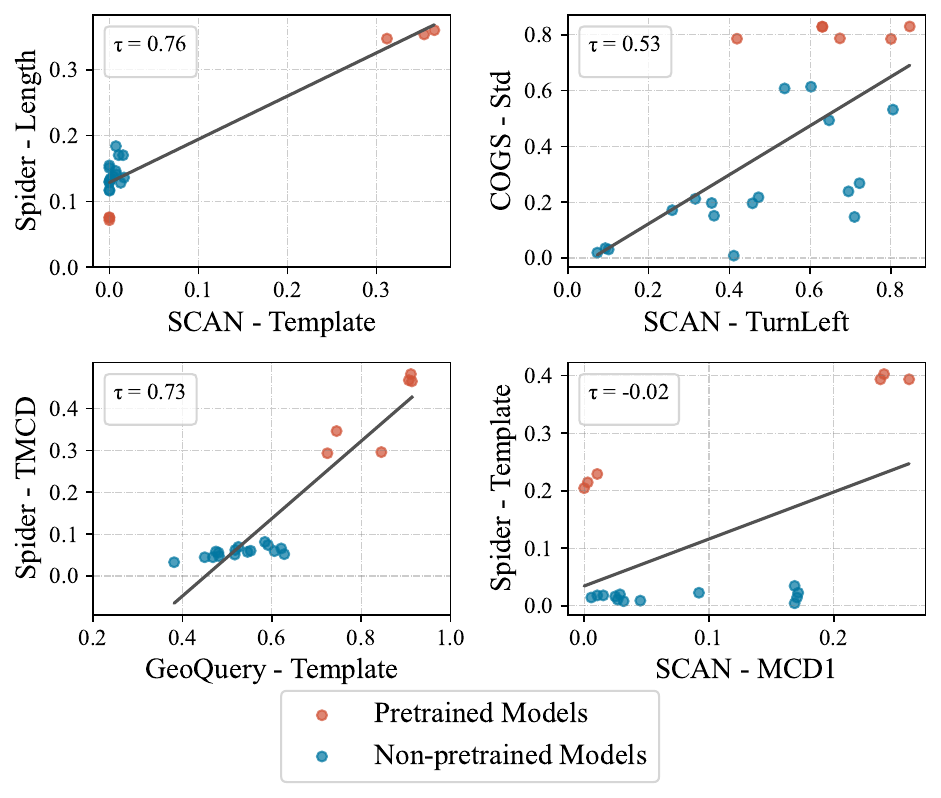}
\caption{Performance of one dataset split versus another. Upper left is an example of high concurrence pair between a synthetic and a natural dataset; upper right is an example of low concurrence within synthetic datasets; lower left is an example of high concurrence within natural datasets; lower right is an example of low concurrence between natural and synthetic datasets.}
\label{fig:perfvperf}
\end{figure}

Why are these concurrence values so low? %
The first hypothesis that we explore is that performance on strongly structured templated data may not correlate with performance on datasets that are authored by humans. To this end, we compute the average concurrence values of three combinations of dataset split pairs, natural-natural, natural-synthetic and synthetic-synthetic, and include an example of each pair type in \cref{fig:perfvperf}. 
We find that splits of natural datasets concur much better than splits of synthetic datasets ($0.54$ v.s.\ $0.22$); the worst is concurrence between synthetic and natural dataset splits ($0.19$).
The same finding can be observed in \cref{fig:perfvconcur}, which we will use later to explore the relationship between concurrence values and performance in \cref{sec:perf:confounding}.

These results are in line with earlier studies that suggested that performance on synthetic compositionality datasets may not transfer to more realistic scenarios \citep{chaabouni-etal-2021-transformers, shaw-etal-2021-compositional}, and underline the point made by \citet{dankers-etal-2022-paradox}, who argue that compositionality should be studied in its natural habitat.
Also the concurrence between dataset splits with naturalistic data is well below the threshold for high concurrence, suggesting that there exist factors beyond dataset creation strategy that can affect how compositionality benchmarks rank modeling approaches.

\subsection{Interpretations of compositionality}

\begin{figure}[t]
\centering
\includegraphics[width=0.85\columnwidth]{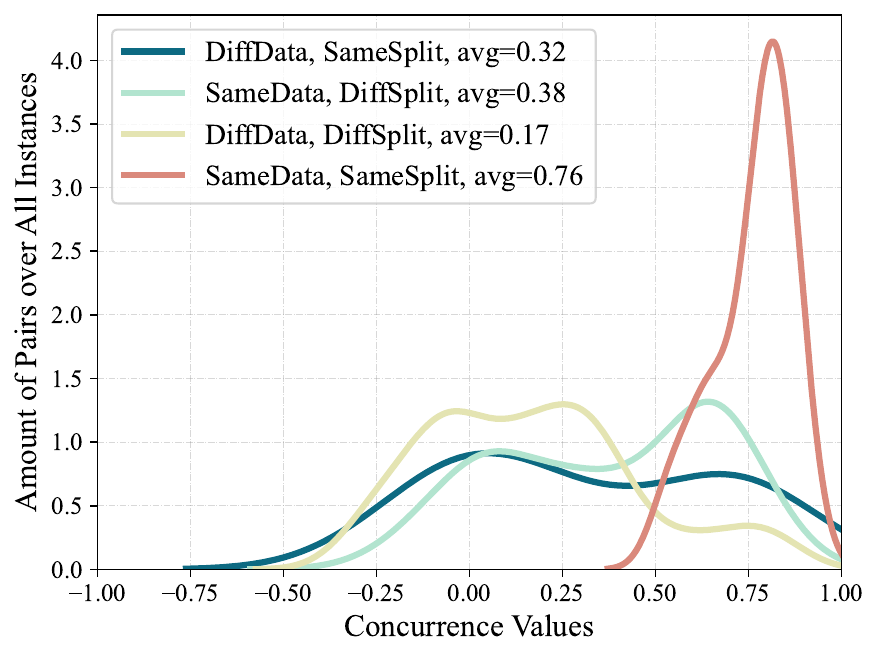}
\caption{Distribution of concurrence values among all dataset splits. The color of the bar indicates whether the splits in the pair share the same dataset origin and/or the same splitting strategy.}
\label{fig:pairtype-bar}
\end{figure}
\label{sec:comp_def}

The next hypothesis that we consider is that concurrence values are low because different dataset splits investigate different types of compositionality \citep[cf.][]{hupkes2020compositionality}.
In compositional evaluation datasets, the interpretation of compositionality is operationalized through its \emph{splitting strategy}.
One splitting strategy may, for instance, define compositional generalization as generalization to longer lengths, whereas another instead focuses on generalization to novel vocabulary items.
These different interpretations of compositionality could potentially require different model capabilities.
Could it be that our concurrence values are low because different splits in fact focus on different types of compositional generalization?

To investigate this, we group the concurrence values by four dataset pair types -- different datasets with the same splitting strategy, the same dataset with different splitting strategies, different datasets with different splitting strategies, and the same dataset with the same splitting strategy -- and plot them in \cref{fig:pairtype-bar}.
Predictably, datasets concur most with themselves (red line).
We also see that which data a splitting approach is applied to is more important than the interpretation of compositionality (cyan and dark blue lines, respectively): concurrence between experiments that share the same source of data averages at $0.38$, whereas different data but the same splitting strategy results in an average concurrence of $0.32$.
However, when both the source of data and splitting strategy are different (yellow line), the concurrence values  shift leftward, suggesting that the data type and splitting strategy pose different kinds of difficulties for the modelling approaches considered.

\newcommand{\phstar}[0]{\phantom{$^*$}}

\begin{table}[t]
\scriptsize
\centering
\begin{tabular}{llllll}
\toprule
\textbf{Dataset A} & \textbf{Dataset B} & \textbf{Concur} & \textbf{Dataset A} & \textbf{Dataset B} & \textbf{Concur} \\ 
\cmidrule(lr){1-3} \cmidrule(lr){4-6}
COGS      & GeoQuery  & \phstar  0.54  & COGS      & SCAN      & \phstar  0.01  \\
COGS      & Spider    & \phstar  0.26 & SCAN      & Spider    & \phstar  0.01   \\
GeoQuery  & Spider    & \phstar  0.23  & GeoQuery  & SCAN      &  - 0.09  \\
 \bottomrule
\end{tabular}
\caption{Concurrence between length splits of datasets.}
\label{tab:length}
\end{table}

\paragraph{Length Generalization}
Because not every dataset in previous work applied all the splitting strategies, we follow-up with a small experiment in a split shared across all datasets: \emph{length generalization} splits.\footnote{As the original COGS dataset did not come with a length generalization split, we generate one ourselves.}
The concurrence values between the different length splits, shown in \cref{tab:length}, are generally low, ranging from $-0.09$ to $0.54$ and averaging at 0.16.
This additional experiment confirms that even when benchmarks maintain the same interpretation of compositionality, there may still be substantial differences in model rankings, depending on the underlying data.

\subsection{The influence of lexical items}
\label{sec:lex_exposure}
\begin{figure}[t]
\centering
\includegraphics[width=0.85\columnwidth]{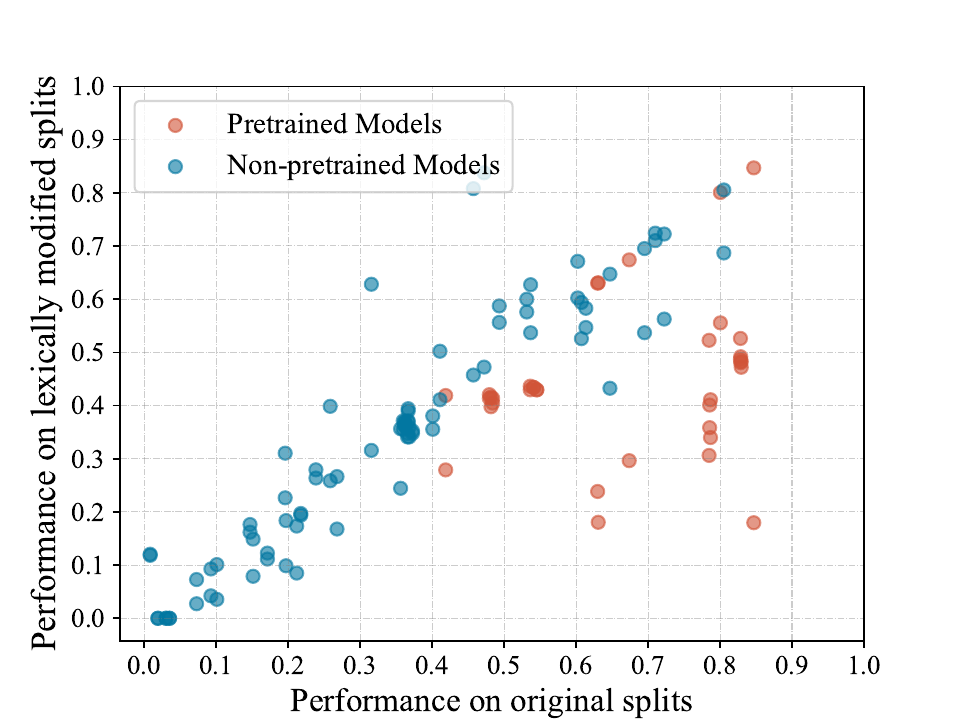}
\caption{Performance of the original split versus the splits with lexical items replaced. Performance of pretrained models decreases when train on the splits with lexical items that are not previously seen in pretraining.}
\label{fig:std_v_lex}
\end{figure}
\begin{table}[t]
\begin{subtable}[h]{\the\columnwidth}
    \centering
\scriptsize
\begin{tabular}{llll}
\toprule
\textbf{Dataset}          & \textbf{Split A}          & \textbf{Split B} & \textbf{Concur} \\ \midrule
\multirow{4}{*}{GeoQuery} & \multirow{2}{*}{Std}      & Std-Rcvcv        & 0.69            \\
                          &                           & Std-Rstr         & 0.54            \\
                          & \multirow{2}{*}{TMCD}     & TMCD-Rstr        & 0.65            \\
                          &                           & TMCD-Rcvcv       & 0.63            \\ \midrule
\multirow{2}{*}{COGS}     & \multirow{2}{*}{Std}      & RandStr          & 0.60            \\
                          &                           & Randcvcv         & 0.59            \\ \midrule
\multirow{2}{*}{SCAN}     & \multirow{2}{*}{TurnLeft} & TurnLeftRcvcv    & 0.29            \\
                          &                           & TurnLeftRStr     & 0.23 \\ \bottomrule       
\end{tabular}
\caption{Concurrence between the original split and lexically-processed splits.}
\label{tab:sub:lex-concur}
\end{subtable}
\hfill
\begin{subtable}[h]{\the\columnwidth}

\begin{adjustbox}{max width=\columnwidth}
\scriptsize

\begin{tabular}{lllll}
\toprule
\textbf{Dataset A} & \textbf{Split A} & \textbf{Dataset B} & \textbf{Split B} & \textbf{Concur} \\ \midrule
COGS     & \textit{Length}     & GeoQuery & \textit{TMCD-Rcvcv} & 0.84 \\
GeoQuery & \textit{Std-Rcvcv}  & GeoQuery & \textit{TMCD-Rcvcv} & 0.83 \\
COGS     & \textit{Std}        & GeoQuery & \textit{TMCD-Rcvcv} & 0.82 \\
GeoQuery & \textit{TMCD-Rstr}  & Spider   & \textit{Template}   & 0.82 \\
GeoQuery & \textit{TMCD-Rcvcv} & Spider   & \textit{Template}   & 0.81 \\
COGS     & \textit{Length}     & GeoQuery & \textit{TMCD-Rstr}  & 0.81 \\
COGS     & \textit{Length}     & GeoQuery & \textit{Std-Rcvcv}  & 0.8  \\
GeoQuery & \textit{Std-Rcvcv}  & GeoQuery & \textit{TMCD-Rstr}  & 0.8  \\
\textbf{GeoQuery}  & \textit{\textbf{TMCD-Rstr}}  & \textbf{Spider}    & \textit{\textbf{TMCD}} & \textbf{0.79}   \\
\textbf{GeoQuery}  & \textit{\textbf{TMCD-Rcvcv}} & \textbf{Spider}    & \textit{\textbf{TMCD}} & \textbf{0.79}   \\
COGS     & \textit{Std}        & GeoQuery & \textit{Std-Rcvcv}  & 0.78 \\
GeoQuery & \textit{Std}        & GeoQuery & \textit{TMCD-Rstr}  & 0.77 \\
GeoQuery & \textit{Std}        & GeoQuery & \textit{TMCD-Rcvcv} & 0.75 \\
COGS     & \textit{Std}        & GeoQuery & \textit{TMCD-Rstr}  & 0.74 \\
GeoQuery & \textit{Template}   & Spider   & \textit{TMCD}       & 0.73 \\
GeoQuery & \textit{Std-Rcvcv}  & Spider   & \textit{Template}   & 0.73 \\
COGS     & \textit{RandStr}    & GeoQuery & \textit{Std-Rstr}   & 0.73 \\
COGS     & \textit{Std}        & GeoQuery & \textit{Std-Rstr}   & 0.72 \\
GeoQuery & \textit{Std-Rstr}   & GeoQuery & \textit{TMCD-Rcvcv} & 0.71 \\
GeoQuery & \textit{Std-Rcvcv}  & Spider   & \textit{TMCD}       & 0.71 \\
COGS     & \textit{Randcvcv}   & GeoQuery & \textit{Std-Rstr}   & 0.7  \\ \bottomrule
\end{tabular}
\end{adjustbox}
\caption{High concurrence values after introducing lexically-processed splits, excluding self-concurrence or concurrence between lexically-processed splits that share the same origin.}
\label{tab:sub:lex-high-concur}
\end{subtable}

\caption{Performance and Concurrence between the lexically-processed splits of datasets.}
\label{tab:lexical}
\end{table}

In \cref{tab:perfStd}, we can see that pretrained models achieve the highest accuracies and in \cref{tab:concurrence} that the highest concurrence values are between two natural datasets. 
In this section, we dive into the differences between pretrained and trained-from-scratch models,  and investigate the extent to which those differences affect the concurrence results.
In particular, we investigate whether the presence of uncontrolled lexical exposure during pretraining may impact the performance of pretrained models, implying their accuracy numbers may not solely reflect their compositional abilities, as suggested by \citet{kim2022uncontrolled}.
Were this to happen, a misalignment in the evaluation between pretrained and non-pretrained models would contribute to variation in the concurrence values, where the performance of pretrained models is overestimated due to lexical exposure in pretraining.

To test for possible effects of lexical exposure, we extend the experiment from \citet{kim2022uncontrolled} -- who conducted it for COGS -- to the TMCD and Std split of GeoQueory, and the TurnLeft split of SCAN\footnote{In both these cases, particular lexical items are purposefully left out of the training set, to be evaluated at test time. If those lexical items were also present in the uncontrolled pretraining corpus, this would thus break the test.}
In both cases, we swap out lexical items with strings of similar length that act as ``wug words'' \citep{berko-1958-child}, or, in other words, previously unattested and therefore meaningless lexical items.
Following \citet{kim2022uncontrolled}, we generate the strings in two ways:
\begin{compactitem}
    \item \textit{Rstr:} We randomly sample lowercase characters from the Latin script with replacements.
    \item \textit{Rcvcv:} We alternately sample a vowel after a consonant from the Latin script.
\end{compactitem}

\noindent We train the models on all modified splits and compute the performance (\cref{fig:std_v_lex}). 
We also compute the concurrence between the original split and the modified split (\cref{tab:sub:lex-concur} and \cref{tab:sub:lex-high-concur}).

In \cref{fig:std_v_lex}, we see that the performance of the pretrained models drops drastically when the lexical items are replaced, while the non-pretrained models' performance does not, confirming the results of \citet{kim2022uncontrolled}.
In addition, the concurrence between the original splits and the modified splits for all datasets is below our set threshold -- albeit higher than other comparisons we have seen before (Table~\ref{tab:sub:lex-concur}) -- implying that replacing lexical items results in yet another new ranking of modeling approaches for compositionality.

We then compute the concurrence between the same set of splits before and after the lexical exposure edits: \emph{within} the group of splits that are selected for the lexical changes, the concurrence values decrease from $0.49$ to $0.41$, while the average concurrence values of these splits with \emph{other} splits that haven't undergone lexical edits slightly increase from $0.25$ to $0.26$ (e.g. concurrence between GeoQuery and Spider TMCD splits increases when GeoQuery TMCD split applies the lexical changes), with many more dataset split pairs surpassing the $\tau=0.7$ bar for high concurrence (\cref{tab:sub:lex-high-concur}).

A closer look explains this apparent contrast: the overall low-concurring dataset SCAN -- which makes up 12.5\% of the lexically edited splits, drags down the concurrence values within that group.
Excluding SCAN, the within-group concurrence values also increase, from $0.63$ to $0.66$.
These results do thus not only confirm that controlling lexical exposure is important when evaluating compositionality in pretrained models, but also further exemplify our earlier finding that compositionality scores -- for neural models -- strongly depend task and dataset. We further analyze the influence of tasks to compositionality results in \cref{app:sec:nacs}.

\subsection{Other confounding factors}\label{sec:confounding_factors}
\label{sec:perf:confounding}
\begin{figure}[t]
\centering
\includegraphics[width=0.85\columnwidth]{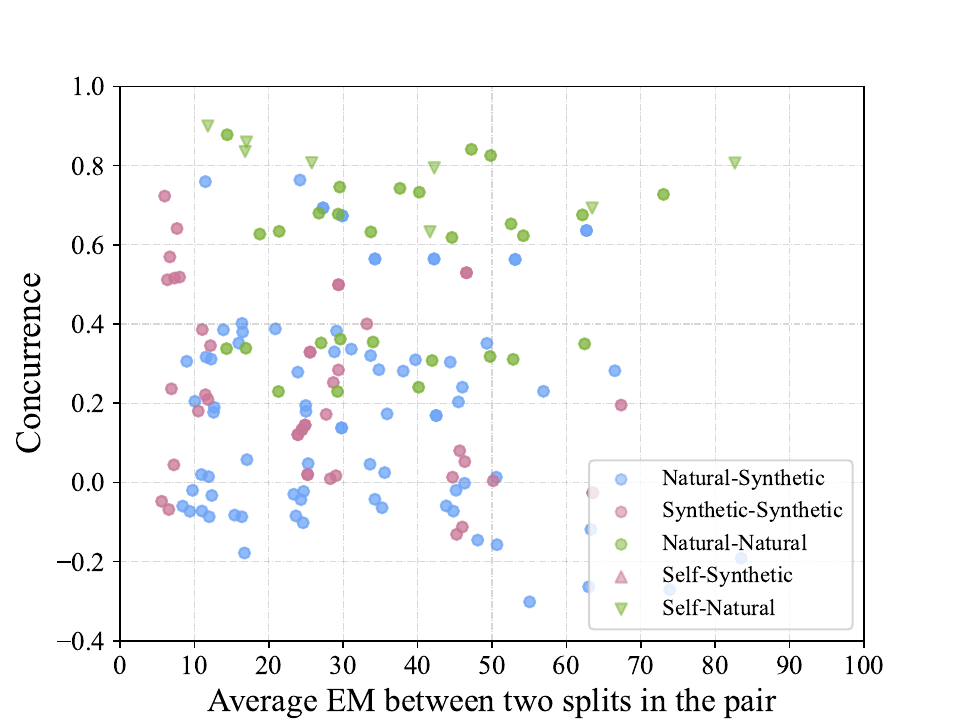}
\caption{Values of concurrences with respect to pairwise averaged performance among the splits shown in Table~\ref{tab:perfStd}. The color of dots indicates the type of split pairs. The triangle-shape dots indicates the values of self-concurrence.}
\label{fig:perfvconcur}
\end{figure}

We have explored a range of factors that may impact the evaluation of compositionality, such as the nature of the underlying data and task, the interpretation of compositionality, and the choice of lexical items.
We wrap up our analysis by verifying that our results are not driven by specific performance scores: we verify that concurrence values are not skewed by datasets for which performances are saturated or close to random.
To assess this, we compute the correlation between the average performance between two datasets and their concurrence, as plotted in \cref{fig:perfvconcur}.
As can be seen, there is no apparent relation between average performance and concurrence: difficult datasets do not concur less or more than easier ones, and dataset saturation (or the opposite: random performance) appears not to impact the results.
A correlation test confirms this visually observed pattern: the Pearson correlation coefficient between performance and concurrence is near zero ($r=0.026$).

\section{Conclusion}
In this paper, we explored how different evaluation choices impact the conclusions drawn from the experiments evaluating compositionality.
Using compositional generalization datasets and models ranging from trained-from-scratch to pretrained, we conduct a series of experiments to understand whether datasets consistently rank models in terms of their generalizability, and we find little consistency. When we perform further analysis to try to better understand this inconsistency, we find that comparing within the training setting (pretrained v.s.\ trained-from-scratch) or data creation type (synthetically generated v. naturally generated) does not increase consistency.
However, better controlling the lexical items can help us draw more consistent conclusions, at least for datasets that share the same notion of compositionality.
We leave the investigation into how task selection might affect evaluation results for compositional generalization to further research.
Overall, our results suggest that to evaluate compositional generalization consistently, clearer definitions of compositionality are needed, as well as %
more careful consideration of evaluation design and more thorough dataset evaluations.

\bibliography{anthology,custom}

\begin{thebibliography}{49}
\expandafter\ifx\csname natexlab\endcsname\relax\def\natexlab#1{#1}\fi

\bibitem[{Bastings et~al.(2018)Bastings, Baroni, Weston, Cho, and Kiela}]{bastings-etal-2018-jump}
Jasmijn Bastings, Marco Baroni, Jason Weston, Kyunghyun Cho, and Douwe Kiela. 2018.
\newblock \href {https://doi.org/10.18653/v1/W18-5407} {Jump to better conclusions: {SCAN} both left and right}.
\newblock In \emph{Proceedings of the 2018 {EMNLP} Workshop {B}lackbox{NLP}: Analyzing and Interpreting Neural Networks for {NLP}}, pages 47--55, Brussels, Belgium. Association for Computational Linguistics.

\bibitem[{Berko(1958)}]{berko-1958-child}
Jean Berko. 1958.
\newblock The child's learning of english morphology.
\newblock \emph{Word}, 14(2-3):150--177.

\bibitem[{Chaabouni et~al.(2021)Chaabouni, Dess{\`\i}, and Kharitonov}]{chaabouni-etal-2021-transformers}
Rahma Chaabouni, Roberto Dess{\`\i}, and Eugene Kharitonov. 2021.
\newblock \href {https://doi.org/10.18653/v1/2021.blackboxnlp-1.9} {Can transformers jump around right in natural language? assessing performance transfer from {SCAN}}.
\newblock In \emph{Proceedings of the Fourth BlackboxNLP Workshop on Analyzing and Interpreting Neural Networks for NLP}, pages 136--148, Punta Cana, Dominican Republic. Association for Computational Linguistics.

\bibitem[{Chowdhery et~al.(2022)Chowdhery, Narang, Devlin, Bosma, Mishra et~al.}]{chowdhery-etal-2022-scaling}
Aakanksha Chowdhery, Sharan Narang, Jacob Devlin, Maarten Bosma, Gaurav Mishra, et~al. 2022.
\newblock \href {https://doi.org/10.48550/arXiv.2204.02311} {{PaLM}: {S}caling language modeling with pathways}.
\newblock \emph{CoRR}, abs/2204.02311.

\bibitem[{Cui et~al.(2022)Cui, Aralikatte, Lent, and Hershcovich}]{cui-etal-2022-compositional}
Ruixiang Cui, Rahul Aralikatte, Heather Lent, and Daniel Hershcovich. 2022.
\newblock \href {https://doi.org/10.1162/tacl_a_00499} {Compositional generalization in multilingual semantic parsing over {W}ikidata}.
\newblock \emph{Transactions of the Association for Computational Linguistics}, 10:937--955.

\bibitem[{Dankers et~al.(2022)Dankers, Bruni, and Hupkes}]{dankers-etal-2022-paradox}
Verna Dankers, Elia Bruni, and Dieuwke Hupkes. 2022.
\newblock \href {https://doi.org/10.18653/v1/2022.acl-long.286} {The paradox of the compositionality of natural language: A neural machine translation case study}.
\newblock In \emph{Proceedings of the 60th Annual Meeting of the Association for Computational Linguistics (Volume 1: Long Papers)}, pages 4154--4175, Dublin, Ireland. Association for Computational Linguistics.

\bibitem[{Finegan-Dollak et~al.(2018)Finegan-Dollak, Kummerfeld, Zhang, Ramanathan, Sadasivam, Zhang, and Radev}]{finegan2018improving}
Catherine Finegan-Dollak, Jonathan~K Kummerfeld, Li~Zhang, Karthik Ramanathan, Sesh Sadasivam, Rui Zhang, and Dragomir Radev. 2018.
\newblock \href {https://aclanthology.org/P18-1033/} {Improving text-to-{SQL} evaluation methodology}.
\newblock In \emph{Proceedings of the 56th Annual Meeting of the Association for Computational Linguistics (Volume 1: Long Papers)}, pages 351--360.

\bibitem[{Hochreiter and Schmidhuber(1997)}]{hochreiter1997long}
Sepp Hochreiter and J{\"u}rgen Schmidhuber. 1997.
\newblock Long short-term memory.
\newblock \emph{Neural computation}, 9(8):1735--1780.

\bibitem[{Hosseini et~al.(2022)Hosseini, Vani, Bahdanau, Sordoni, and Courville}]{hosseini-etal-2022-compositional}
Arian Hosseini, Ankit Vani, Dzmitry Bahdanau, Alessandro Sordoni, and Aaron Courville. 2022.
\newblock \href {https://aclanthology.org/2022.blackboxnlp-1.22} {On the compositional generalization gap of in-context learning}.
\newblock In \emph{Proceedings of the Fifth BlackboxNLP Workshop on Analyzing and Interpreting Neural Networks for NLP}, pages 272--280, Abu Dhabi, United Arab Emirates (Hybrid). Association for Computational Linguistics.

\bibitem[{Hupkes et~al.(2020)Hupkes, Dankers, Mul, and Bruni}]{hupkes2020compositionality}
Dieuwke Hupkes, Verna Dankers, Mathijs Mul, and Elia Bruni. 2020.
\newblock Compositionality decomposed: How do neural networks generalise?
\newblock \emph{Journal of Artificial Intelligence Research}, 67:757--795.

\bibitem[{Hupkes et~al.(2023)Hupkes, Giulianelli, Dankers, Artetxe, Elazar, Pimentel, Christodoulopoulos, Lasri, Saphra, Sinclair, Ulmer, Schottmann, Batsuren, Sun, Sinha, Khalatbari, Ryskina, Frieske, Cotterell, and Jin}]{Hupkes2023}
Dieuwke Hupkes, Mario Giulianelli, Verna Dankers, Mikel Artetxe, Yanai Elazar, Tiago Pimentel, Christos Christodoulopoulos, Karim Lasri, Naomi Saphra, Arabella Sinclair, Dennis Ulmer, Florian Schottmann, Khuyagbaatar Batsuren, Kaiser Sun, Koustuv Sinha, Leila Khalatbari, Maria Ryskina, Rita Frieske, Ryan Cotterell, and Zhijing Jin. 2023.
\newblock \href {https://doi.org/10.1038/s42256-023-00729-y} {A taxonomy and review of generalization research in nlp}.
\newblock \emph{Nature Machine Intelligence}, 5(10):1161--1174.

\bibitem[{Jacobs and Wallach(2021)}]{jacobs-wallach-2021-measurement}
Abigail~Z Jacobs and Hanna Wallach. 2021.
\newblock Measurement and fairness.
\newblock In \emph{Proceedings of the 2021 ACM conference on fairness, accountability, and transparency}, pages 375--385.

\bibitem[{Keysers et~al.(2020)Keysers, Sch{\"a}rli, Scales, Buisman, Furrer, Kashubin, Momchev, Sinopalnikov, Stafiniak, Tihon, Tsarkov, Wang, van Zee, and Bousquet}]{keysers2020measuring}
Daniel Keysers, Nathanael Sch{\"a}rli, Nathan Scales, Hylke Buisman, Daniel Furrer, Sergii Kashubin, Nikola Momchev, Danila Sinopalnikov, Lukasz Stafiniak, Tibor Tihon, Dmitry Tsarkov, Xiao Wang, Marc van Zee, and Olivier Bousquet. 2020.
\newblock \href {https://openreview.net/forum?id=SygcCnNKwr} {Measuring compositional generalization: A comprehensive method on realistic data}.
\newblock In \emph{International Conference on Learning Representations}.

\bibitem[{Keysers et~al.(2019)Keysers, Sch{\"a}rli, Scales, Buisman, Furrer, Kashubin, Momchev, Sinopalnikov, Stafiniak, Tihon et~al.}]{keysers2019measuring}
Daniel Keysers, Nathanael Sch{\"a}rli, Nathan Scales, Hylke Buisman, Daniel Furrer, Sergii Kashubin, Nikola Momchev, Danila Sinopalnikov, Lukasz Stafiniak, Tibor Tihon, et~al. 2019.
\newblock \href {https://openreview.net/pdf?id=SygcCnNKwr} {Measuring compositional generalization: A comprehensive method on realistic data}.
\newblock In \emph{International Conference on Learning Representations}.

\bibitem[{Kim and Linzen(2020)}]{kim-linzen-2020-cogs}
Najoung Kim and Tal Linzen. 2020.
\newblock \href {https://doi.org/10.18653/v1/2020.emnlp-main.731} {{COGS}: A compositional generalization challenge based on semantic interpretation}.
\newblock In \emph{Proceedings of the 2020 Conference on Empirical Methods in Natural Language Processing (EMNLP)}, pages 9087--9105, Online. Association for Computational Linguistics.

\bibitem[{Kim et~al.(2022)Kim, Linzen, and Smolensky}]{kim2022uncontrolled}
Najoung Kim, Tal Linzen, and Paul Smolensky. 2022.
\newblock Uncontrolled lexical exposure leads to overestimation of compositional generalization in pretrained models.
\newblock \emph{arXiv preprint arXiv:2212.10769}.

\bibitem[{Kim(2021)}]{kim2021sequencetosequence}
Yoon Kim. 2021.
\newblock \href {https://openreview.net/forum?id=pbfAgoc_l2w} {Sequence-to-sequence learning with latent neural grammars}.
\newblock In \emph{Advances in Neural Information Processing Systems}.

\bibitem[{Klein et~al.(2017)Klein, Kim, Deng, Senellart, and Rush}]{klein-etal-2017-opennmt}
Guillaume Klein, Yoon Kim, Yuntian Deng, Jean Senellart, and Alexander Rush. 2017.
\newblock \href {https://aclanthology.org/P17-4012} {{O}pen{NMT}: Open-source toolkit for neural machine translation}.
\newblock In \emph{Proceedings of {ACL} 2017, System Demonstrations}, pages 67--72, Vancouver, Canada. Association for Computational Linguistics.

\bibitem[{Lake and Baroni(2018)}]{lake2018generalization}
Brenden Lake and Marco Baroni. 2018.
\newblock Generalization without systematicity: On the compositional skills of sequence-to-sequence recurrent networks.
\newblock In \emph{International conference on machine learning}, pages 2873--2882. PMLR.

\bibitem[{Lewis et~al.(2020)Lewis, Liu, Goyal, Ghazvininejad, Mohamed, Levy, Stoyanov, and Zettlemoyer}]{lewis-etal-2020-bart}
Mike Lewis, Yinhan Liu, Naman Goyal, Marjan Ghazvininejad, Abdelrahman Mohamed, Omer Levy, Veselin Stoyanov, and Luke Zettlemoyer. 2020.
\newblock \href {https://doi.org/10.18653/v1/2020.acl-main.703} {{BART}: Denoising sequence-to-sequence pre-training for natural language generation, translation, and comprehension}.
\newblock In \emph{Proceedings of the 58th Annual Meeting of the Association for Computational Linguistics}, pages 7871--7880, Online. Association for Computational Linguistics.

\bibitem[{Li et~al.(2021)Li, Yin, Chen, and Zhang}]{li-etal-2021-compositional}
Yafu Li, Yongjing Yin, Yulong Chen, and Yue Zhang. 2021.
\newblock \href {https://doi.org/10.18653/v1/2021.acl-long.368} {On compositional generalization of neural machine translation}.
\newblock In \emph{Proceedings of the 59th Annual Meeting of the Association for Computational Linguistics and the 11th International Joint Conference on Natural Language Processing (Volume 1: Long Papers)}, pages 4767--4780, Online. Association for Computational Linguistics.

\bibitem[{Liska et~al.(2018)Liska, Kruszewski, and Baroni}]{liska2018memorize}
Adam Liska, Germ{\'{a}}n Kruszewski, and Marco Baroni. 2018.
\newblock \href {http://arxiv.org/abs/1802.06467} {Memorize or generalize? searching for a compositional {RNN} in a haystack}.
\newblock \emph{CoRR}, abs/1802.06467.

\bibitem[{Liu et~al.(2021)Liu, Lee, Jia, and Liang}]{liu2021question}
Nelson~F Liu, Tony Lee, Robin Jia, and Percy Liang. 2021.
\newblock Do question answering modeling improvements hold across benchmarks?
\newblock \emph{arXiv preprint arXiv:2102.01065}.

\bibitem[{Loula et~al.(2018)Loula, Baroni, and Lake}]{loula-etal-2018-rearranging}
Jo{\~a}o Loula, Marco Baroni, and Brenden Lake. 2018.
\newblock \href {https://doi.org/10.18653/v1/W18-5413} {Rearranging the familiar: Testing compositional generalization in recurrent networks}.
\newblock In \emph{Proceedings of the 2018 {EMNLP} Workshop {B}lackbox{NLP}: Analyzing and Interpreting Neural Networks for {NLP}}, pages 108--114, Brussels, Belgium. Association for Computational Linguistics.

\bibitem[{Luo et~al.(2022)Luo, Xu, and Xiong}]{luo-etal-2022-cogtaskonomy}
Yifei Luo, Minghui Xu, and Deyi Xiong. 2022.
\newblock \href {https://doi.org/10.18653/v1/2022.acl-long.64} {{C}og{T}askonomy: Cognitively inspired task taxonomy is beneficial to transfer learning in {NLP}}.
\newblock In \emph{Proceedings of the 60th Annual Meeting of the Association for Computational Linguistics (Volume 1: Long Papers)}, pages 904--920, Dublin, Ireland. Association for Computational Linguistics.

\bibitem[{Oren et~al.(2021)Oren, Herzig, and Berant}]{oren-etal-2021-finding}
Inbar Oren, Jonathan Herzig, and Jonathan Berant. 2021.
\newblock \href {https://doi.org/10.18653/v1/2021.emnlp-main.843} {Finding needles in a haystack: Sampling structurally-diverse training sets from synthetic data for compositional generalization}.
\newblock In \emph{Proceedings of the 2021 Conference on Empirical Methods in Natural Language Processing}, pages 10793--10809, Online and Punta Cana, Dominican Republic. Association for Computational Linguistics.

\bibitem[{Orhan(2021)}]{orhan2021compositional}
A~Emin Orhan. 2021.
\newblock Compositional generalization in semantic parsing with pretrained transformers.
\newblock \emph{arXiv preprint arXiv:2109.15101}.

\bibitem[{Padmakumar et~al.(2022)Padmakumar, Lausen, Ballesteros, Zha, He, and Karypis}]{padmakumar-etal-2022-exploring}
Vishakh Padmakumar, Leonard Lausen, Miguel Ballesteros, Sheng Zha, He~He, and George Karypis. 2022.
\newblock \href {https://doi.org/10.18653/v1/2022.naacl-main.183} {Exploring the role of task transferability in large-scale multi-task learning}.
\newblock In \emph{Proceedings of the 2022 Conference of the North American Chapter of the Association for Computational Linguistics: Human Language Technologies}, pages 2542--2550, Seattle, United States. Association for Computational Linguistics.

\bibitem[{Patel et~al.(2022)Patel, Bhattamishra, Blunsom, and Goyal}]{patel-etal-2022-revisiting}
Arkil Patel, Satwik Bhattamishra, Phil Blunsom, and Navin Goyal. 2022.
\newblock \href {https://doi.org/10.18653/v1/2022.acl-short.46} {Revisiting the compositional generalization abilities of neural sequence models}.
\newblock In \emph{Proceedings of the 60th Annual Meeting of the Association for Computational Linguistics (Volume 2: Short Papers)}, pages 424--434, Dublin, Ireland. Association for Computational Linguistics.

\bibitem[{Raffel et~al.(2020)Raffel, Shazeer, Roberts, Lee, Narang, Matena, Zhou, Li, Liu et~al.}]{raffel2020exploring}
Colin Raffel, Noam Shazeer, Adam Roberts, Katherine Lee, Sharan Narang, Michael Matena, Yanqi Zhou, Wei Li, Peter~J Liu, et~al. 2020.
\newblock Exploring the limits of transfer learning with a unified text-to-text transformer.
\newblock \emph{J. Mach. Learn. Res.}, 21(140):1--67.

\bibitem[{Raunak et~al.(2019)Raunak, Kumar, Metze, and Callan}]{raunak2019compositionality}
Vikas Raunak, Vaibhav Kumar, Florian Metze, and Jaimie Callan. 2019.
\newblock \href {https://vaibhav4595.github.io/files/compo.pdf} {On compositionality in neural machine translation}.
\newblock In \emph{NeurIPS 2019 Context and Compositionality in Biological and Artificial Neural Systems Workshop}.

\bibitem[{Reddy et~al.(2017)Reddy, T{\"a}ckstr{\"o}m, Petrov, Steedman, and Lapata}]{reddy-etal-2017-universal}
Siva Reddy, Oscar T{\"a}ckstr{\"o}m, Slav Petrov, Mark Steedman, and Mirella Lapata. 2017.
\newblock \href {https://doi.org/10.18653/v1/D17-1009} {Universal semantic parsing}.
\newblock In \emph{Proceedings of the 2017 Conference on Empirical Methods in Natural Language Processing}, pages 89--101, Copenhagen, Denmark. Association for Computational Linguistics.

\bibitem[{Ruis et~al.(2020)Ruis, Andreas, Baroni, Bouchacourt, and Lake}]{ruis-etal-2020-gscan}
Laura Ruis, Jacob Andreas, Marco Baroni, Diane Bouchacourt, and Brenden~M. Lake. 2020.
\newblock A benchmark for systematic generalization in grounded language understanding.
\newblock In \emph{Advances in Neural Information Processing Systems 33: Annual Conference on Neural Information Processing Systems 2020, NeurIPS 2020, December 6-12, 2020, virtual}.

\bibitem[{Shaw et~al.(2021)Shaw, Chang, Pasupat, and Toutanova}]{shaw-etal-2021-compositional}
Peter Shaw, Ming-Wei Chang, Panupong Pasupat, and Kristina Toutanova. 2021.
\newblock \href {https://doi.org/10.18653/v1/2021.acl-long.75} {Compositional generalization and natural language variation: Can a semantic parsing approach handle both?}
\newblock In \emph{Proceedings of the 59th Annual Meeting of the Association for Computational Linguistics and the 11th International Joint Conference on Natural Language Processing (Volume 1: Long Papers)}, pages 922--938, Online. Association for Computational Linguistics.

\bibitem[{Sun et~al.(2023{\natexlab{a}})Sun, Qi, Zhang, Liu, Wang, and Huang}]{sun2022tokenization}
Kaiser Sun, Peng Qi, Yuhao Zhang, Lan Liu, William~Yang Wang, and Zhiheng Huang. 2023{\natexlab{a}}.
\newblock Tokenization consistency matters for generative models on extractive {NLP} tasks.
\newblock In \emph{Findings of the Association for Computational Linguistics: EMNLP 2023}. Association for Computational Linguistics.

\bibitem[{Sun et~al.(2023{\natexlab{b}})Sun, Williams, and Hupkes}]{sun-etal-2023-resp}
Kaiser Sun, Adina Williams, and Dieuwke Hupkes. 2023{\natexlab{b}}.
\newblock {A} replication study of compositional generalization works on semantic parsing.
\newblock \emph{ReScience C}, 9(2):44.

\bibitem[{Tang and Mooney(2001)}]{tang2001using}
Lappoon~R Tang and Raymond~J Mooney. 2001.
\newblock Using multiple clause constructors in inductive logic programming for semantic parsing.
\newblock In \emph{European Conference on Machine Learning}, pages 466--477. Springer.

\bibitem[{Thrush et~al.(2022)Thrush, Jiang, Bartolo, Singh, Williams, Kiela, and Ross}]{thrush-etal-2022-winoground}
Tristan Thrush, Ryan Jiang, Max Bartolo, Amanpreet Singh, Adina Williams, Douwe Kiela, and Candace Ross. 2022.
\newblock Winoground: Probing vision and language models for visio-linguistic compositionality.
\newblock In \emph{Proceedings of the IEEE/CVF Conference on Computer Vision and Pattern Recognition}, pages 5238--5248.

\bibitem[{Valvoda et~al.(2022)Valvoda, Saphra, Rawski, Williams, and Cotterell}]{valvoda-etal-2022-benchmarking}
Josef Valvoda, Naomi Saphra, Jonathan Rawski, Adina Williams, and Ryan Cotterell. 2022.
\newblock \href {https://aclanthology.org/2022.coling-1.525} {Benchmarking compositionality with formal languages}.
\newblock In \emph{Proceedings of the 29th International Conference on Computational Linguistics}, pages 6007--6018, Gyeongju, Republic of Korea. International Committee on Computational Linguistics.

\bibitem[{Vaswani et~al.(2017)Vaswani, Shazeer, Parmar, Uszkoreit, Jones, Gomez, Kaiser, and Polosukhin}]{vaswani2017attention}
Ashish Vaswani, Noam Shazeer, Niki Parmar, Jakob Uszkoreit, Llion Jones, Aidan~N Gomez, {\L}ukasz Kaiser, and Illia Polosukhin. 2017.
\newblock Attention is all you need.
\newblock \emph{Advances in neural information processing systems}, 30.

\bibitem[{Vu et~al.(2020)Vu, Wang, Munkhdalai, Sordoni, Trischler, Mattarella-Micke, Maji, and Iyyer}]{vu-etal-2020-exploring}
Tu~Vu, Tong Wang, Tsendsuren Munkhdalai, Alessandro Sordoni, Adam Trischler, Andrew Mattarella-Micke, Subhransu Maji, and Mohit Iyyer. 2020.
\newblock \href {https://doi.org/10.18653/v1/2020.emnlp-main.635} {Exploring and predicting transferability across {NLP} tasks}.
\newblock In \emph{Proceedings of the 2020 Conference on Empirical Methods in Natural Language Processing (EMNLP)}, pages 7882--7926, Online. Association for Computational Linguistics.

\bibitem[{Wang et~al.(2022)Wang, Titov, Andreas, and Kim}]{wang-etal-2022-hierarchical-phrase}
Bailin Wang, Ivan Titov, Jacob Andreas, and Yoon Kim. 2022.
\newblock \href {https://aclanthology.org/2022.emnlp-main.563} {Hierarchical phrase-based sequence-to-sequence learning}.
\newblock In \emph{Proceedings of the 2022 Conference on Empirical Methods in Natural Language Processing}, pages 8211--8229, Abu Dhabi, United Arab Emirates. Association for Computational Linguistics.

\bibitem[{Weber et~al.(2021)Weber, Jumelet, Bruni, and Hupkes}]{weber-etal-2021-language}
Lucas Weber, Jaap Jumelet, Elia Bruni, and Dieuwke Hupkes. 2021.
\newblock \href {https://doi.org/10.18653/v1/2021.eacl-main.176} {Language modelling as a multi-task problem}.
\newblock In \emph{Proceedings of the 16th Conference of the European Chapter of the Association for Computational Linguistics: Main Volume}, pages 2049--2060, Online. Association for Computational Linguistics.

\bibitem[{Westen and Rosenthal(2003)}]{westen2003quantifying}
Drew Westen and Robert Rosenthal. 2003.
\newblock Quantifying construct validity: two simple measures.
\newblock \emph{Journal of personality and social psychology}, 84(3):608.

\bibitem[{Wu(1997)}]{wu-1997-stochastic}
Dekai Wu. 1997.
\newblock \href {https://aclanthology.org/J97-3002} {Stochastic inversion transduction grammars and bilingual parsing of parallel corpora}.
\newblock \emph{Computational Linguistics}, 23(3):377--403.

\bibitem[{Yao and Koller(2022)}]{yao-koller-2022-structural}
Yuekun Yao and Alexander Koller. 2022.
\newblock \href {https://aclanthology.org/2022.emnlp-main.337} {Structural generalization is hard for sequence-to-sequence models}.
\newblock In \emph{Proceedings of the 2022 Conference on Empirical Methods in Natural Language Processing}, pages 5048--5062, Abu Dhabi, United Arab Emirates. Association for Computational Linguistics.

\bibitem[{Ye et~al.(2021)Ye, Lin, and Ren}]{ye-etal-2021-crossfit}
Qinyuan Ye, Bill~Yuchen Lin, and Xiang Ren. 2021.
\newblock \href {https://doi.org/10.18653/v1/2021.emnlp-main.572} {{C}ross{F}it: A few-shot learning challenge for cross-task generalization in {NLP}}.
\newblock In \emph{Proceedings of the 2021 Conference on Empirical Methods in Natural Language Processing}, pages 7163--7189, Online and Punta Cana, Dominican Republic. Association for Computational Linguistics.

\bibitem[{Yu et~al.(2018)Yu, Zhang, Yang, Yasunaga, Wang, Li, Ma, Li, Yao, Roman, Zhang, and Radev}]{yu-etal-2018-Spider}
Tao Yu, Rui Zhang, Kai Yang, Michihiro Yasunaga, Dongxu Wang, Zifan Li, James Ma, Irene Li, Qingning Yao, Shanelle Roman, Zilin Zhang, and Dragomir Radev. 2018.
\newblock \href {https://doi.org/10.18653/v1/D18-1425} {{S}pider: A large-scale human-labeled dataset for complex and cross-domain semantic parsing and text-to-{SQL} task}.
\newblock In \emph{Proceedings of the 2018 Conference on Empirical Methods in Natural Language Processing}, pages 3911--3921, Brussels, Belgium. Association for Computational Linguistics.

\bibitem[{Zelle and Mooney(1996)}]{zelle1996learning}
John~M Zelle and Raymond~J Mooney. 1996.
\newblock Learning to parse database queries using inductive logic programming.
\newblock In \emph{Proceedings of the national conference on artificial intelligence}, pages 1050--1055.

\end{thebibliography}
\bibliographystyle{acl_natbib}

\clearpage

\appendix

\section{Dataset examples}
\label{app:sec:data}
For convenience, we include a brief description with examples of all datasets we consider in our experiments in \cref{tab:examples}.
The description of each split and the number of instances in each dataset split is shown in \cref{app:tab:splitintro} and \cref{tab:num_instances}.

\paragraph{SCAN} Consisting of a set of commands and the corresponding action sequences, SCAN \cite{lake2018generalization} is one of the most popular synthetic datasets to study compositional generalization.
The model is given commands like \texttt{jump left} and is expected to predict action sequences like \texttt{LTURN JUMP}.
We include the \textit{simple}, \textit{length}, \textit{add primitive}, \textit{template} splits from \citet{lake2018generalization}.
In addition to original SCAN splits, we also use maximum compound divergence (MCD) splits of SCAN proposed by \citet{keysers2020measuring}.

\paragraph{COGS} 
\citet{kim-linzen-2020-cogs} introduce COGS, a synthetic semantic parsing dataset generated by a rule-based approach, which covers a larger variety of grammar rules than SCAN does.
The inputs in COGS are English sentences, generated by a probabilistic context-free grammar. 
The corresponding output, which is the semantic interpretation of the input, is annotated with the logical formalism in \citet{reddy-etal-2017-universal}.
COGS includes a randomly sampled test set and an out-of-distribution compositional generalization set.

\paragraph{GeoQuery} GeoQuery \cite{tang2001using, zelle1996learning} is a text-to-QL dataset containing naturalistic examples.
We use the four compositional generalization splits defined on this dataset by \citet{shaw-etal-2021-compositional}: 
We use the splits in \citet{shaw-etal-2021-compositional}, in which all entity mentions are converted with placeholders and use Functional Query Language (FunQL) as the target representation.
\textit{random/standard}, \textit{length}, \textit{template}, and \textit{Target Maximum Compound Divergence (TMCD)}.
The TMCD split is an extension of MCD splits in SCAN, with the capability to be applied to non-synthetic datasets.

\paragraph{Spider} Spider \cite{yu-etal-2018-Spider} is originally designed for cross-domain semantic parsing, and targets a challenging kind of generalization, generalization to new database schemata, using different databases for the training and test set. 
It also uses SQL for a more complex syntax. 
We use the compositional generalization splits for Spider defined by \citet{shaw-etal-2021-compositional}, which match their splits for GeoQuery: 
\textit{random/standard}, \textit{length}, \textit{template}, and \textit{TMCD}.
In the same paper, \citet{shaw-etal-2021-compositional} split Spider into the same four splits as GeoQuery and adopt a setting where databases are shared between train and test examples so that the dataset splits can be dedicated to evaluating compositional generalization.

\begin{table*}[ht]
\scriptsize
\centering
\begin{adjustbox}{max width=\textwidth}
\begin{tabular}{p{4em}p{5em}p{0.8\textwidth}}
\toprule
\multirow{2}{*}{\textbf{COGS}} & \textbf{Input:} & Mila liked that the cake was offered to Emma . \\
& \textbf{Output:} & \texttt{* cake ( x \_ 4 ) ; like . agent ( x \_ 1 , Mila ) AND like . ccomp ( x \_ 1 , x \_ 6 ) AND offer . theme ( x \_ 6 , x \_ 4 ) AND offer . recipient ( x \_ 6 , Emma )}
\\ \midrule
\multirow{2}{*}{\textbf{SCAN}}  &  \textbf{Input:} & turn left after jump twice \\
 & \textbf{Output:} & \texttt{I\_JUMP I\_JUMP I\_TURN\_LEFT}
\\ \midrule
\multirow{2}{*}{\textbf{NACS}}  &  \textbf{Input:} & run thrice after jump around left \\
 & \textbf{Output:} & \texttt{I\_TURN\_LEFT I\_JUMP I\_TURN\_LEFT I\_JUMP I\_TURN\_LEFT I\_JUMP I\_TURN\_LEFT I\_JUMP I\_RUN I\_RUN I\_RUN}
\\ \midrule
\multirow{2}{*}{\textbf{GeoQuery}}  &  \textbf{Input:} & how much population does m0 have \\
&  \textbf{Output:} & \texttt{answer ( intersection ( river , loc\_2 ( m0 ) ) )} \\ \midrule
\multirow{2}{*}{\textbf{Spider}}  &  \textbf{Input:} & flight\_1: what is the average distance and price for all flights from la?  \\
&  \textbf{Output:} & \texttt{select avg(distance) , avg(price) from flight where origin = "los angeles"} \\
\bottomrule
\end{tabular}
\end{adjustbox}
\caption{Examples of instances in each dataset used in our experiments.}\label{tab:examples}

\end{table*}
\begin{table*}[ht]
\small
\centering
\begin{adjustbox}{max width=\textwidth}
\begin{tabular}{lll}
\toprule
\textbf{Split} & \textbf{Dataset} & \textbf{Description}                                       \\
\midrule
\textit{random/standard/simple}              & COGS, SCAN, GeoQuery, Spider              & Split the dataset randomly. \\
\textit{length} & COGS, SCAN, GeoQuery, Spider & Split the dataset according to the input length. \\
\textit{template} & SCAN, GeoQuery, Spider & Split the dataset based on a given string template. \\
\textit{TurnLeft} & SCAN & Compositional commands of \texttt{TurnLeft} are isolated in training set. \\
\textit{Jump} & SCAN & Compositional commands of \texttt{Jump} are isolated in training set. \\
\textit{MCD} & SCAN & Split according to maximum compound divergence. \\
\textit{TMCD} & GeoQuery, Spider & Natural counterpart of MCD, split the data based on target MCD. \\
\textit{Gen} & COGS & \begin{tabular}[c]{@{}l@{}}Not a splitting strategy, but a collection of specially generated samples \\ designed to test 21 cases of generalization in COGS. \end{tabular}
 \\

\bottomrule
\end{tabular}
\end{adjustbox}
\caption{Summary of each split and their designated dataset we use.}
\label{app:tab:splitintro}
\end{table*}
\begin{table*}[ht]
\centering
\scriptsize
\begin{adjustbox}{max width=\textwidth}
\begin{tabular}{lllllll}
\toprule
\textbf{Dataset} & \textbf{Split}          & \textbf{Train} & \textbf{Validation} & \textbf{Test}  & \textbf{Overall} \\
\midrule
\multirow{4}{*}{COGS}     & no\_mod                  & 24155 & 3000       & 3000  & 21000 & 51155   \\
                          & random\_cvcv             & 24155 & 3000       & 3000  & 21000 & 51155   \\
                          & random\_str              & 24155 & 3000       & 3000  & 21000 & 51155   \\
                          & length                   & 24156 & -          & 23999 & -     & 48155   \\
                          \midrule
\multirow{4}{*}{GeoQuery} & standard                 & 600   & -          & 280   & -     & 880     \\
                          & length                   & 440   & -          & 440   & -     & 880     \\
                          & template                 & 441   & -          & 439   & -     & 880     \\
                          & tmcd                     & 440   & -          & 440   & -     & 880     \\
                          \midrule
\multirow{11}{*}{SCAN}    & simple                   & 16728 & -          & 4182  & -     & 20910   \\
                          & length                   & 16990 & -          & 3920  & -     & 20910   \\
                          & mcd1                     & 8365  & 1045       & 1045  & -     & 10455   \\
                          & mcd2                     & 8365  & 1045       & 1045  & -     & 10455   \\
                          & mcd3                     & 8365  & 1045       & 1045  & -     & 10455   \\
                          & addprim\_jump            & 14670 & -          & 7706  & -     & 22376   \\
                          & addprim\_turn\_left      & 21890 & -          & 1208  & -     & 23098   \\
                          & jump\_random\_cvcv       & 14670 & -          & 7706  & -     & 22376   \\
                          & jump\_random\_str        & 14670 & -          & 7706  & -     & 22376   \\
                          & turn\_left\_random\_cvcv & 21890 & -          & 1208  & -     & 23098   \\
                          & turn\_left\_random\_str  & 21890 & -          & 1208  & -     & 23098   \\
                          \midrule
\multirow{4}{*}{Spider}   & random                   & 3282  & -          & 1094  & -     & 4376    \\
                          & length                   & 3282  & -          & 1094  & -     & 4376    \\
                          & template                 & 3280  & -          & 1096  & -     & 4376    \\
                          & tmcd                     & 3282  & -          & 1094  & -     & 4376    \\
\bottomrule
\end{tabular}
\end{adjustbox}
\caption{Number of instances for each dataset in each optimization split.}
\label{tab:num_instances}
\end{table*}

\section{License of Artifacts}
We include the licenses and intended usage of artifacts used in this work in \cref{app:tab:licenses}.
\begin{table*}[ht]
\scriptsize
\centering
\begin{adjustbox}{max width=\textwidth}
\begin{tabular}{lll}
\toprule
\textbf{Artifact} & \textbf{License} & \textbf{Intended Usage}                                       \\
\midrule
COGS              & MIT              & A dataset focuses on compositional generalization \\
SCAN & BSD & A dataset focuses on compositional generalization. \\
GeoQuery & ODC-BY 1.0 license & A database query datasets for U.S. geography. \\
Spider & CC BY-SA 4.0 & A cross-domain semantic parsing and text-to-SQL dataset. \\ 
NACS & CC-NC & A dataset focuses on compositional generalization. \\ \midrule
Neural-BTG              & MIT       & \begin{tabular}[c]{@{}l@{}}A neural transducer for sequence-to-sequence tasks.\end{tabular}   \\
\begin{tabular}[c]{@{}l@{}}LSTM, Transformer \\ (OpenNMT-py \cite{klein-etal-2017-opennmt}) \end{tabular}           & MIT       &  Models for  sequence-to-sequence tasks. \\
T5              & Apache-2.0       & \begin{tabular}[c]{@{}l@{}}A pre-trained model for sequence-to-sequence tasks.\end{tabular}   \\
BART              & Apache-2.0       & \begin{tabular}[c]{@{}l@{}}A pre-trained model for sequence-to-sequence tasks.\end{tabular}   \\
\bottomrule
\end{tabular}
\end{adjustbox}
\caption{License and intended usage for the artifacts we used.}
\label{app:tab:licenses}
\end{table*}

\section{Hyperparameters}
\label{app:hp_tuning}
For the models and dataset combinations that have already been trained by prior works, we adopt the same set of hyperparameters. 
For the remaining combinations, we tune the hyperparameters on a random split of the original dataset, with 90\% data in the training set and 10\% data in the test set.
We describe the final hyperparamters below.

For T5 with GEOQUERY and SPIDER, we follow the same hyperparameter setup as \citealp{shaw-etal-2021-compositional}. For LSTM and Transformer with COGS, we follow the same hyperparameter setup as in \citealp{kim-linzen-2020-cogs}. For T5 with COGS, we follow the training strategy from \cite{orhan2021compositional}.

For other datasets, we tune the learning rate of T5 and BART in [$10^{-5}, 10^{-4}, 10^{-3}$]. We tune the dropout rate in [$0.0$, $0.1$, $0.5$] and layers in [$1$, $2$] for LSTMs; dropout rate in [$0.0$, $0.1$, $0.5$] and layers in [$2$, $4$, $8$] for Transformer. For BTG, we tune the vocabulary size between 200 and 800, as well as the learning rate in [$1.0 \times 10^{-4}, 3.0 \times 10^{-4}$].

\section{Evaluation: Variants of Exact Match Accuracy}
\label{app:sec:EM}
\begin{table}[H]
\scriptsize
\centering
\begin{tabular}{llrrr}
\toprule
\textbf{Dataset}      & \textbf{Split}    & \multicolumn{1}{l}{\textbf{T5}} & \multicolumn{1}{l}{\textbf{BART}} & \multicolumn{1}{l}{\textbf{BTG}} \\ \midrule
\multirow{7}{*}{COGS} & \textit{Std-Test} & 99.7                            & 0.0                               & 0.0                              \\
                          & \textit{Std-Gen}    & 82.9 & 0.0  & 0.0  \\
                          & \textit{Rcvcv-Test} & 99.7 & 0.0  & 0.0  \\
                          & \textit{Rstr-Test}  & 99.8 & 0.0  & 0.0  \\
                          & \textit{Rcvcv-Gen}  & 50.0 & 0.0  & 0.0  \\
                          & \textit{Rstr-Gen}   & 48.0 & 0.0  & 0.0  \\
                          & \textit{Length}     & 37.9 & 0.0  & 0.0  \\ \midrule
\multirow{4}{*}{Spider}   & Rand                & 60.1 & 26.2 & 32.4 \\
                          & Template            & 34.9 & 18.1 & 1.8  \\
                          & TMCD                & 38.3 & 23.5 & 4.9  \\
                          & Length              & 33.9 & 6.1  & 11.9 \\ \midrule
\multirow{8}{*}{GeoQuery} & Std                 & 77.1 & 0.0  & 0.0  \\
                          & Std-Rcvcv           & 74.3 & 0.0  & 0.0  \\
                          & Std-Rstr            & 73.5 & 0.0  & 0.0  \\
                          & Template            & 76.5 & 0.0  & 0.0  \\
                          & Length              & 39.5 & 0.0  & 0.0  \\
                          & TMCD                & 40.7 & 0.0  & 0.0  \\
                          & TMCD-Rcvcv          & 31.6 & 0.0  & 0.0  \\
                          & TMCD-Rstr           & 31.4 & 0.0  & 0.0  \\ \bottomrule
\end{tabular}
\caption{Percentage difference between raw EM implementation and EM implementation that ignore harmless space (space-lenient EM - raw EM). SCAN and NACS are omitted because models do not have this issue on them. LSTMs do not display this issue; the difference for Transformer is under 0.1\% for each datset.}
\label{tab:perf_difference}
\end{table}
\begin{table*}[ht]
\centering
\scriptsize
\begin{adjustbox}{max width=\textwidth}
\begin{tabular}{p{7em}p{0.8\textwidth}}
\toprule
\textbf{Input:} & Zoe thought that a hippo cleaned . \\
\textbf{Output:} & think\hl{ }. agent ( x \_ 1\hl{ }, Zoe ) AND think\hl{ }. ccomp ( x \_ 1\hl{ }, x \_ 5 ) AND hippo ( x \_ 4 ) AND clean\hl{ }. agent ( x \_ 5\hl{ }, x \_ 4 ) \\
\textbf{Prediction:} & think. agent ( x \_ 1, Zoe ) AND think. ccomp ( x \_ 1, x \_ 5 ) AND hippo ( x \_ 4 ) AND clean. agent ( x \_ 5, x \_ 4 ) \\
\bottomrule
\end{tabular}
\end{adjustbox}
\caption{Examples of instance where the model is only mistaken on the space.}
\label{tab:space-example}

\end{table*}
The most intuitive implementation of exact match accuracy is directly comparing the output text string with the gold sequence, without any post-processing. 
However, we found this to be unnecessarily strict for some models, such as T5, which does not have the ``<" symbol, which appears in a large number of instances, in the vocabulary and required post-processing to replace the UNK tokens with ``<".
In addition, although the location of space should not change the correctness of a prediction for our evaluated datasets, often incorrect spaces led to wrong evaluation when direct text comparison is used. \cref{tab:space-example} shows an example of such an instance.
With the leniency on spaces, T5's exact match value changed from zero accuracy on a whole dataset (COGS) to performing among the best on all datasets (\cref{tab:perf_difference}); this is likely due to the tokenization of special tokens with space, as noted in \citet{sun2022tokenization}.
\section{Spider performance}
\begin{table}[H]
\centering
\scriptsize
\begin{tabular}{p{0.1\columnwidth}p{0.09\columnwidth}p{0.09\columnwidth}p{0.09\columnwidth}p{0.09\columnwidth}p{0.09\columnwidth}p{0.09\columnwidth}}
\toprule
\textbf{Split} &
   \multicolumn{1}{l}{\textbf{\begin{tabular}[c]{@{}l@{}}LSTM\\  Uni\end{tabular}}} &
  \multicolumn{1}{l}{\textbf{\begin{tabular}[c]{@{}l@{}}LSTM\\  Bi\end{tabular}}} &
  \multicolumn{1}{l}{\textbf{\begin{tabular}[c]{@{}l@{}}Trans-\\ former\end{tabular}}} &
  \multicolumn{1}{l}{\textbf{T5}} &
  \multicolumn{1}{l}{\textbf{BART}} &
  \multicolumn{1}{l}{\textbf{BTG}} \\ \midrule
\textit{Rand}     & 0.0 & 0.0 & 0.0 & 77.8 & 34.8 & 46.2 \\
\textit{Template} & 1.4 & 2.7 & 3.2 & 52.5 & 25.5 & 3.5  \\
\textit{TMCD}     & 0.1 & 0.1 & 0.1 & 57.6 & 37.9 & 6.9  \\
\textit{Length}   & 0.9 & 0.6 & 0.3 & 44.4 & 9.0  & 16.5 \\ \bottomrule
\end{tabular}
\caption{Model exact-match accuracy with Spider EM. A large amount of output of LSTM and Transformer are deemed as invalid SQL due to special tokens.}
\label{tab:app:spider-perf}
\end{table}
\label{app:sec:spider}
The official release of Spider \cite{yu-etal-2018-Spider} uses a different variant of exact match accuracy, which is more lenient than the version we used.
We include a table of model performance on splits of Spider, evaluated with the official Spider metric in \cref{tab:app:spider-perf}.

\section{The influence of task similarity}
\label{app:sec:nacs}
As briefly mentioned in \cref{sec:lex_exposure}, task formulation can be another factor that affects the agreement between datasets.
To understand the effect of task similarity on the conclusion obtained from compositionality benchmarks, we add in the NACS dataset \cite{bastings-etal-2018-jump} for existing experiments, as all three datasets except for SCAN are semantic parsing tasks, while SCAN falls under a navigation task.
NACS is introduced as a dataset that is similar to SCAN but requires mapping actions back to the original commands, and it is thus more complex for models compared to SCAN and will not allow simple models to gain unintended high performance.
We train models on NACS with the same hyperparameter tuning and training strategy as in \cref{sec:Methods}, compute the concurrence between NACS and other datasets, and look at the effect of different splitting strategy between SCAN and NACS. 
The results are discussed below.

\subsection{Overall Performance and Concurrence}
\begin{figure}[t]
\centering
\includegraphics[width=\the\columnwidth]{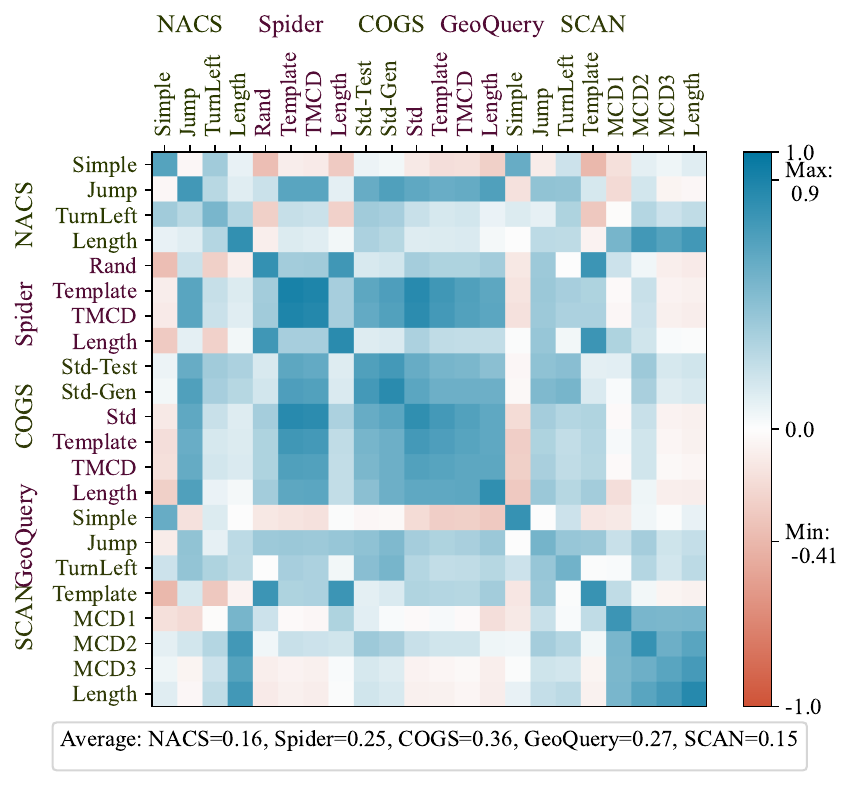}
    \caption{Distribution of concurrence values between each dataset and split pairs.}
\label{app:fig:nacs-default_conf_matrix}
\end{figure}
\begin{figure}[t]
\centering
\includegraphics[width=0.85\columnwidth]{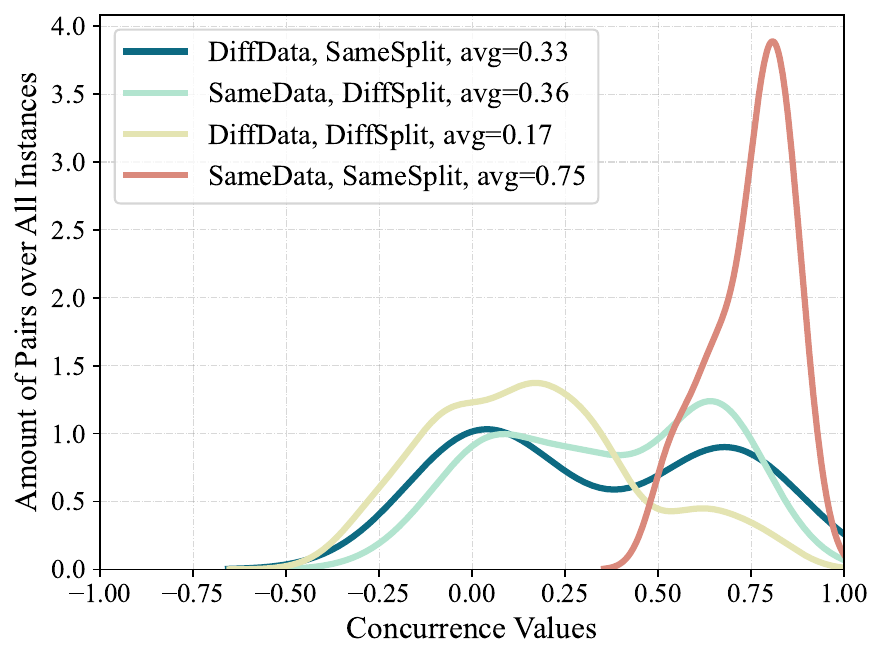}
\caption{Distribution of concurrence values among all dataset splits. The color of the bar indicates whether the splits in the pair share the same dataset origin and/or the same splitting strategy.}
\label{app:fig:nacs-pairtype-bar}
\end{figure}
\begin{table}[t]
\centering
\scriptsize
\begin{tabular}{llllc}
\toprule
\textbf{Dataset A} & \textbf{Dataset B} & \textbf{Split A} & \textbf{Split B} & \textbf{Concur} \\ \midrule
Spider   & Spider   & \textit{Template} & \textit{TMCD}     & 0.88 \\
GeoQuery & Spider   & \textit{Std}      & \textit{Template} & 0.84 \\
GeoQuery & Spider   & \textit{Std}      & \textit{TMCD}     & 0.83 \\
SCAN     & Spider   & \textit{Template} & \textit{Rand}     & 0.76 \\
SCAN     & Spider   & \textit{Template} & \textit{Length}   & 0.76 \\
Spider   & Spider   & \textit{Rand}     & \textit{Length}   & 0.75 \\
GeoQuery & Spider   & \textit{Template} & \textit{Template} & 0.74 \\
SCAN     & NACS     & \textit{MCD2}     & \textit{Length}   & 0.74 \\
GeoQuery & Spider   & \textit{Template} & \textit{TMCD}     & 0.73 \\
SCAN     & NACS     & \textit{Length}   & \textit{Length}   & 0.73 \\
GeoQuery & GeoQuery & \textit{Std}      & \textit{Template} & 0.73 \\
SCAN     & SCAN     & \textit{Length}   & \textit{MCD3}     & 0.72 \\
\bottomrule
\end{tabular}
\caption{High concurrence values ($\geq 0.7$) among all pairs of dataset splits, excluding self-concurrence.}
\label{app:tab:NACS:concurrence}
\end{table}
The overall performance and concurrence including NACS are shown in \cref{app:tab:all-perf} and \cref{app:fig:nacs-default_conf_matrix}.
The concurrence values between NACS and SCAN is surprisingly low compared to the concurrence values between NACS and other datasets, with the \textit{length} split being the only exception, suggesting that even when the underlying tasks are the same, the datasets may provide very different model rankings.
In terms of the distribution of concurrence values by type of data split pairs (\cref{app:fig:nacs-pairtype-bar}), the conclusion in \cref{sec:comp_def} persists: the source of the dataset matters more than the interpretation of compositonality (splitting strategy).

\subsection{Length Split of NACS}

\begin{table}[t]
\scriptsize
\centering
\begin{tabular}{llllll}
\toprule
\textbf{Dataset A} & \textbf{Dataset B} & \textbf{Concur} & \textbf{Dataset A} & \textbf{Dataset B} & \textbf{Concur} \\ 
\cmidrule(lr){1-3} \cmidrule(lr){4-6}

SCAN      & NACS      & \phstar 0.73   & GeoQuery & NACS   & \phstar 0.08  \\
COGS      & GeoQuery  & \phstar 0.54   & Spider   & NACS   & \phstar 0.04  \\
COGS      & Spider    & \phstar 0.26   & SCAN     & Spider & \phstar 0.01  \\
COGS      & NACS      & \phstar 0.24   & COGS     & SCAN   & \phstar 0.01  \\
GeoQuery  & Spider    & \phstar 0.23   & GeoQuery & SCAN   & \phstar -0.09 \\
 \bottomrule
\end{tabular}
\caption{Concurrence between length splits of datasets.}
\label{app:tab:nacs-length}
\end{table}
Out of the four splits of NACS, the \textit{length} split is the only split that results in a high concurrence with tsplits of SCAN (\cref{app:fig:nacs-default_conf_matrix}).
The \textit{length} split of SCAN and NACS is also the only length splits pair that exceed the boundary set for high concurrence (\cref{app:tab:nacs-length}).
It is likely because that both \textit{length} split of NACS and the splits that it has high concurrence with are extremely difficult split that many models fail on.

\section{Performance and concurrence across all setups}
\label{app:sec:all-concur}

\begin{table*}[ht]
\centering
\scriptsize
\begin{adjustbox}{max width=0.9\textwidth}
\begin{tabular}{clrlrlrlrlrlrlr}
\toprule
\textbf{Dataset} &
  \textbf{Split} &
  \multicolumn{2}{c}{\textbf{LSTM Uni}} &
  \multicolumn{2}{c}{\textbf{LSTM Bi}} &
  \multicolumn{2}{c}{\textbf{Transformer}} &
  \multicolumn{2}{c}{\textbf{T5}} &
  \multicolumn{2}{c}{\textbf{BART}} &
  \multicolumn{2}{c}{\textbf{BTG}} &
  \multicolumn{1}{c}{\textbf{Avg}} \\ \midrule
\multirow{7}{*}{COGS}   & \textit{Std-Test}      & 99.3  & ±.0  & 99.1  & ±.01 & 99.5  & ±.0  & 99.7 & ±.0  & 99.7  & ±.0  & 68.8 & ±.01 & 94.3 \\
                        & \textit{Rcvcv-Test}    & 99.4  & ±.0  & 99.1  & ±.0  & 99.5  & ±.0  & 99.7 & ±.0  & 99.7  & ±.0  & 68.1 & ±.0  & 94.2 \\
                        & \textit{Rstr-Test}     & 99.4  & ±.0  & 99.0  & ±.01 & 99.6  & ±.0  & 99.8 & ±.0  & 99.7  & ±.0  & 68.4 & ±.0  & 94.3 \\
                        & \textit{Std-Gen}       & 21.3  & ±.05 & 14.8  & ±.08 & 56.1  & ±.06 & 82.9 & ±.0  & 78.6  & ±.0  & 2.8  & ±.01 & 42.8 \\
                        & \textit{Rcvcv-Gen}     & 22.6  & ±.04 & 10.1  & ±.02 & 57.6  & ±.02 & 50.0 & ±.02 & 44.5  & ±.07 & 0.0  & ±.0  & 30.8 \\
                        & \textit{Rstr-Gen}      & 22.3  & ±.07 & 14.7  & ±.03 & 56.6  & ±.03 & 48.0 & ±.01 & 33.5  & ±.03 & 0.0  & ±.0  & 29.2 \\
                        & \textit{Length}        & 20.7  & ±.01 & 24.9  & ±.01 & 28.7  & ±.02 & 37.9 & ±.0  & 34.1  & ±.01 & 20.5 & ±.0  & 27.8 \\ \midrule
\multirow{10}{*}{SCAN}  & \textit{Simple}        & 99.9  & ±.0  & 99.9  & ±.0  & 100.0 & ±.0  & 94.9 & ±.01 & 99.1  & ±.01 & 12.3 & ±.01 & 84.4 \\
                        & \textit{Jump}          & 0.4   & ±.01 & 0.0   & ±.0  & 0.1   & ±.0  & 95.0 & ±.01 & 0.4   & ±.01 & 0.0  & ±.0  & 16.0 \\
                        & \textit{Template}      & 0.2   & ±.0  & 0.3   & ±.01 & 1.1   & ±.0  & 34.3 & ±.03 & 0.0   & ±.0  & 0.9  & ±.01 & 6.1  \\
                        & \textit{MCD1}          & 5.9   & ±.06 & 12.2  & ±.07 & 1.1   & ±.0  & 24.6 & ±.01 & 0.4   & ±.01 & 1.8  & ±.01 & 7.7  \\
                        & \textit{MCD2}          & 6.7   & ±.03 & 5.8   & ±.03 & 1.2   & ±.0  & 34.1 & ±.01 & 1.6   & ±.0  & 0.5  & ±.0  & 8.3  \\
                        & \textit{MCD3}          & 8.7   & ±.04 & 7.8   & ±.02 & 0.7   & ±.0  & 11.1 & ±.01 & 1.2   & ±.01 & 0.8  & ±.01 & 5.0  \\
                        & \textit{Length}        & 15.3  & ±.04 & 11.8  & ±.01 & 0.0   & ±.0  & 14.1 & ±.01 & 0.7   & ±.01 & 0.0  & ±.0  & 7.0  \\
                        & \textit{TurnLeft}      & 61.1  & ±.13 & 34.1  & ±.06 & 64.8  & ±.11 & 70.3 & ±.12 & 63.1  & ±.19 & 8.9  & ±.01 & 50.4 \\
                        & \textit{TurnLeftRcvcv} & 69.4  & ±.14 & 42.8  & ±.14 & 60.4  & ±.12 & 20.0 & ±.03 & 37.7  & ±.15 & 3.5  & ±.01 & 39.0 \\
                        & \textit{TurnLeftRStr}  & 59.0  & ±.18 & 43.5  & ±.1  & 61.9  & ±.1  & 17.7 & ±.02 & 23.9  & ±.17 & 2.4  & ±.0  & 34.7 \\ \midrule
\multirow{4}{*}{NACS}   & \textit{Simple}        & 100.0 & ±.0  & 100.0 & ±.0  & 100.0 & ±.0  & 94.6 & ±.0  & 100.0 & ±.0  & 6.1  & ±.01 & 83.5 \\
                        & \textit{Jump}          & 0.1   & ±.0  & 0.2   & ±.0  & 0.2   & ±.0  & 95.8 & ±.01 & 67.6  & ±.04 & 0.0  & ±.0  & 27.3 \\
                        & \textit{TurnLeft}      & 63.3  & ±.12 & 62.0  & ±.13 & 54.4  & ±.11 & 64.9 & ±.04 & 82.4  & ±.13 & 9.2  & ±.01 & 56.0 \\
                        & \textit{Length}        & 12.7  & ±.02 & 13.2  & ±.01 & 0.0   & ±.0  & 14.3 & ±.0  & 9.3   & ±.02 & 0.0  & ±.0  & 8.2  \\ \midrule
\multirow{4}{*}{Spider} & \textit{Rand}          & 33.4  & ±.02 & 36.9  & ±.01 & 42.5  & ±.01 & 68.0 & ±.0  & 32.7  & ±.01 & 40.1 & ±.01 & 42.3 \\
                        & \textit{Template}      & 1.0   & ±.0  & 2.2   & ±.01 & 4.6   & ±.0  & 39.6 & ±.01 & 21.6  & ±.01 & 1.9  & ±.0  & 11.8 \\
                        & \textit{TMCD}          & 4.6   & ±.01 & 6.0   & ±.01 & 7.5   & ±.01 & 47.2 & ±.01 & 31.2  & ±.03 & 5.5  & ±.0  & 17.0 \\
                        & \textit{Length}        & 12.7  & ±.01 & 14.0  & ±.01 & 17.5  & ±.01 & 35.4 & ±.01 & 7.4   & ±.0  & 14.0 & ±.01 & 16.8 \\ \midrule
\multirow{8}{*}{GeoQuery} &
  \textit{Std} &
  74.0 &
  ±.06 &
  78.9 &
  ±.04 &
  82.3 &
  ±.02 &
  92.5 &
  ±.01 &
  89.2 &
  ±.01 &
  79.0 &
  ±.01 &
  82.6 \\
                        & \textit{Std-Rcvcv}     & 76.7  & ±.03 & 78.9  & ±.02 & 80.5  & ±.01 & 89.4 & ±.0  & 84.2  & ±.0  & 69.0 & ±.03 & 79.8 \\
                        & \textit{Std-Rstr}      & 77.1  & ±.01 & 78.6  & ±.02 & 82.7  & ±.01 & 88.8 & ±.01 & 79.9  & ±.0  & 65.8 & ±.01 & 78.8 \\
                        & \textit{Template}      & 46.5  & ±.06 & 55.9  & ±.07 & 56.7  & ±.04 & 91.0 & ±.0  & 77.1  & ±.06 & 53.5 & ±.06 & 63.5 \\
                        & \textit{Length}        & 18.5  & ±.03 & 16.2  & ±.02 & 22.0  & ±.01 & 41.1 & ±.01 & 36.1  & ±.01 & 20.7 & ±.02 & 25.8 \\
                        & \textit{TMCD}          & 35.8  & ±.02 & 37.1  & ±.02 & 37.9  & ±.01 & 54.1 & ±.0  & 48.2  & ±.0  & 36.9 & ±.0  & 41.7 \\
                        & \textit{TMCD-Rcvcv}    & 35.9  & ±.01 & 36.7  & ±.01 & 37.5  & ±.0  & 43.3 & ±.0  & 40.8  & ±.01 & 34.3 & ±.0  & 38.1 \\
                        & \textit{TMCD-Rstr}     & 35.5  & ±.01 & 37.7  & ±.01 & 37.6  & ±.0  & 43.1 & ±.0  & 41.4  & ±.0  & 35.3 & ±.01 & 38.4     \\
                        \bottomrule
\end{tabular}
\end{adjustbox}
    \caption{Model exact-match accuracy on datasets averaged across random seeds, with standard deviation.}%
\label{app:tab:all-perf}
\end{table*}
\begin{figure*}[ht]
\centering
\includegraphics[width=0.9\textwidth]{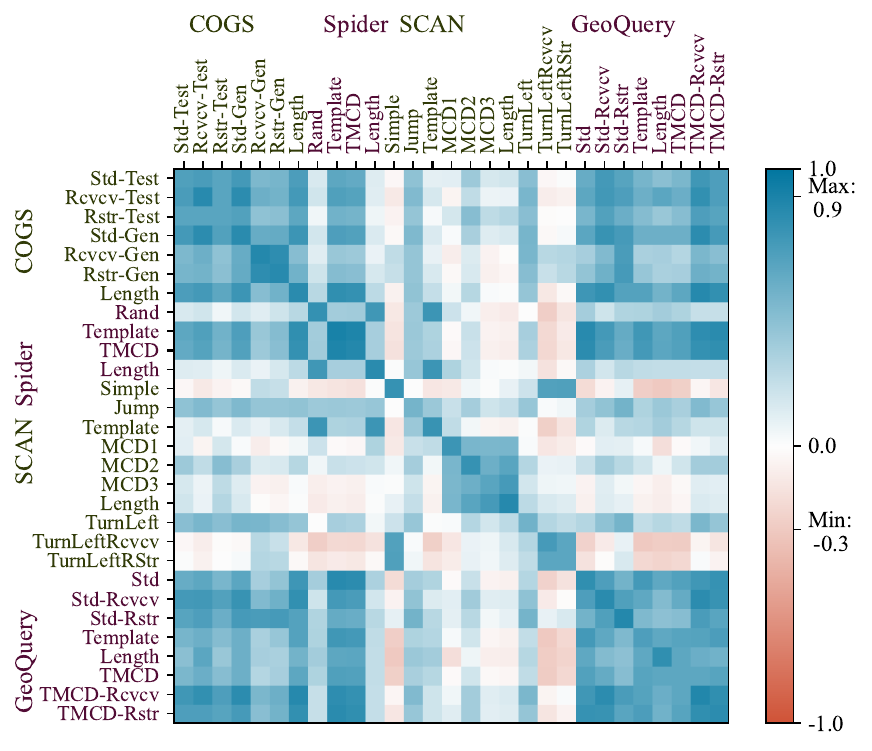}
\caption{Distribution of concurrence values between each dataset and split pairs.}
\label{fig:app:conf_matrix}
\end{figure*}
The performance of all models on all the curated splits for each dataset is shown in \cref{app:tab:all-perf}.
The concurrence between all datasets and split pairs in this work is shown in \cref{fig:app:conf_matrix} and the exact values are included in \cref{tab:app:allconur}.
\begin{table*}[ht]
\centering
\scriptsize
\begin{adjustbox}{max width=\textwidth}
\begin{tabular}{lllllllllc}
\toprule
\textbf{Dataset A} &
  \textbf{Dataset B} &
  \textbf{Split B} &
  \textbf{Split A} &
  \textbf{Concur} &
  \textbf{Dataset A} &
  \textbf{Dataset B} &
  \textbf{Split A} &
  \textbf{Split B} &
  \textbf{Concur} \\ \midrule
Spider   & Spider   & \textit{TMCD}          & \textit{Template}      & 0.88 & COGS     & GeoQuery & \textit{RandStr}    & \textit{TMCD-Rstr}     & 0.54 \\
COGS     & GeoQuery & \textit{TMCD-Rcvcv}    & \textit{Length}        & 0.84 & GeoQuery & SCAN     & \textit{Std-Rstr}   & \textit{TurnLeft}      & 0.54 \\
GeoQuery & Spider   & \textit{Template}      & \textit{Std}           & 0.84 & COGS     & SCAN     & \textit{Std}        & \textit{TurnLeft}      & 0.53 \\
GeoQuery & Spider   & \textit{Template}      & \textit{TMCD-Rstr}     & 0.84 & COGS     & SCAN     & \textit{Randcvcv}   & \textit{TurnLeft}      & 0.52 \\
GeoQuery & GeoQuery & \textit{TMCD-Rcvcv}    & \textit{Std-Rcvcv}     & 0.83 & SCAN     & SCAN     & \textit{MCD1}       & \textit{MCD2}          & 0.52 \\
GeoQuery & Spider   & \textit{TMCD}          & \textit{Std}           & 0.83 & SCAN     & SCAN     & \textit{Length}     & \textit{MCD1}          & 0.52 \\
COGS     & GeoQuery & \textit{TMCD-Rcvcv}    & \textit{Std}           & 0.82 & COGS     & GeoQuery & \textit{Randcvcv}   & \textit{TMCD-Rcvcv}    & 0.51 \\
COGS     & COGS     & \textit{RandStr}       & \textit{Randcvcv}      & 0.82 & GeoQuery & SCAN     & \textit{TMCD-Rcvcv} & \textit{TurnLeft}      & 0.51 \\
GeoQuery & Spider   & \textit{Template}      & \textit{TMCD-Rcvcv}    & 0.81 & SCAN     & SCAN     & \textit{MCD1}       & \textit{MCD3}          & 0.51 \\
COGS     & Spider   & \textit{Template}      & \textit{Length}        & 0.81 & COGS     & SCAN     & \textit{Std}        & \textit{Jump}          & 0.5  \\
GeoQuery & Spider   & \textit{TMCD}          & \textit{TMCD-Rstr}     & 0.81 & GeoQuery & GeoQuery & \textit{Std-Rstr}   & \textit{Template}      & 0.5  \\
GeoQuery & GeoQuery & \textit{TMCD-Rstr}     & \textit{TMCD-Rcvcv}    & 0.81 & GeoQuery & SCAN     & \textit{TMCD-Rcvcv} & \textit{Jump}          & 0.49 \\
COGS     & GeoQuery & \textit{TMCD-Rstr}     & \textit{Length}        & 0.8  & COGS     & GeoQuery & \textit{Randcvcv}   & \textit{Std-Rcvcv}     & 0.49 \\
COGS     & GeoQuery & \textit{Std-Rcvcv}     & \textit{Length}        & 0.8  & GeoQuery & GeoQuery & \textit{Std-Rcvcv}  & \textit{Length}        & 0.48 \\
COGS     & Spider   & \textit{TMCD}          & \textit{Length}        & 0.79 & COGS     & Spider   & \textit{RandStr}    & \textit{Template}      & 0.47 \\
GeoQuery & Spider   & \textit{TMCD}          & \textit{TMCD-Rcvcv}    & 0.79 & COGS     & SCAN     & \textit{RandStr}    & \textit{TurnLeft}      & 0.47 \\
GeoQuery & GeoQuery & \textit{TMCD-Rstr}     & \textit{Std}           & 0.79 & COGS     & COGS     & \textit{Randcvcv}   & \textit{Length}        & 0.47 \\
GeoQuery &
  GeoQuery &
  \textit{TMCD-Rstr} &
  \textit{Std-Rcvcv} &
  0.78 &
  GeoQuery &
  GeoQuery &
  \textit{Std-Rstr} &
  \textit{TMCD} &
  0.46 \\
COGS     & GeoQuery & \textit{Std-Rcvcv}     & \textit{Std}           & 0.78 & COGS     & GeoQuery & \textit{Randcvcv}   & \textit{TMCD-Rstr}     & 0.46 \\
COGS     & COGS     & \textit{Length}        & \textit{Std}           & 0.76 & COGS     & Spider   & \textit{RandStr}    & \textit{TMCD}          & 0.46 \\
SCAN     & Spider   & \textit{Rand}          & \textit{Template}      & 0.76 & GeoQuery & GeoQuery & \textit{Std-Rstr}   & \textit{Length}        & 0.44 \\
SCAN     & Spider   & \textit{Length}        & \textit{Template}      & 0.76 & GeoQuery & SCAN     & \textit{Std-Rcvcv}  & \textit{TurnLeft}      & 0.43 \\
COGS     & GeoQuery & \textit{Std}           & \textit{Length}        & 0.75 & COGS     & SCAN     & \textit{Length}     & \textit{Jump}          & 0.43 \\
GeoQuery & GeoQuery & \textit{TMCD-Rcvcv}    & \textit{Std}           & 0.75 & GeoQuery & SCAN     & \textit{Std-Rcvcv}  & \textit{Jump}          & 0.42 \\
Spider   & Spider   & \textit{Length}        & \textit{Rand}          & 0.75 & COGS     & GeoQuery & \textit{RandStr}    & \textit{Std}           & 0.42 \\
GeoQuery & Spider   & \textit{Template}      & \textit{Template}      & 0.74 & COGS     & SCAN     & \textit{Randcvcv}   & \textit{Jump}          & 0.41 \\
GeoQuery & Spider   & \textit{TMCD}          & \textit{Template}      & 0.73 & GeoQuery & SCAN     & \textit{TMCD-Rstr}  & \textit{TurnLeft}      & 0.41 \\
GeoQuery & Spider   & \textit{Template}      & \textit{Std-Rcvcv}     & 0.73 & COGS     & SCAN     & \textit{Length}     & \textit{TurnLeft}      & 0.41 \\
GeoQuery & GeoQuery & \textit{Template}      & \textit{Std}           & 0.73 & COGS     & SCAN     & \textit{RandStr}    & \textit{Jump}          & 0.41 \\
COGS     & GeoQuery & \textit{Std-Rstr}      & \textit{RandStr}       & 0.73 & COGS     & GeoQuery & \textit{RandStr}    & \textit{Template}      & 0.4  \\
COGS     & GeoQuery & \textit{TMCD-Rstr}     & \textit{Std}           & 0.72 & GeoQuery & SCAN     & \textit{TMCD-Rstr}  & \textit{Jump}          & 0.4  \\
SCAN     & SCAN     & \textit{MCD3}          & \textit{Length}        & 0.72 & SCAN     & Spider   & \textit{Jump}       & \textit{Length}        & 0.4  \\
COGS     & GeoQuery & \textit{Std-Rstr}      & \textit{Std}           & 0.72 & SCAN     & SCAN     & \textit{Jump}       & \textit{TurnLeft}      & 0.4  \\
GeoQuery & GeoQuery & \textit{TMCD-Rcvcv}    & \textit{Std-Rstr}      & 0.71 & COGS     & Spider   & \textit{Randcvcv}   & \textit{Template}      & 0.39 \\
GeoQuery & Spider   & \textit{TMCD}          & \textit{Std-Rcvcv}     & 0.71 & GeoQuery & SCAN     & \textit{Length}     & \textit{Jump}          & 0.39 \\
COGS     & GeoQuery & \textit{Std-Rstr}      & \textit{Randcvcv}      & 0.7  & SCAN     & SCAN     & \textit{Jump}       & \textit{Template}      & 0.39 \\
GeoQuery & GeoQuery & \textit{TMCD-Rstr}     & \textit{Template}      & 0.7  & SCAN     & Spider   & \textit{Jump}       & \textit{Template}      & 0.39 \\
COGS     & Spider   & \textit{Template}      & \textit{Std}           & 0.69 & SCAN     & Spider   & \textit{Jump}       & \textit{Rand}          & 0.38 \\
GeoQuery & GeoQuery & \textit{Std-Rcvcv}     & \textit{Std}           & 0.69 & SCAN     & Spider   & \textit{Jump}       & \textit{TMCD}          & 0.38 \\
SCAN     & SCAN     & \textit{TurnLeftRStr}  & \textit{Simple}        & 0.68 & COGS     & Spider   & \textit{Randcvcv}   & \textit{TMCD}          & 0.38 \\
GeoQuery & GeoQuery & \textit{Std-Rstr}      & \textit{Std-Rcvcv}     & 0.68 & GeoQuery & SCAN     & \textit{Std-Rcvcv}  & \textit{MCD2}          & 0.37 \\
GeoQuery & Spider   & \textit{Template}      & \textit{TMCD}          & 0.68 & Spider   & Spider   & \textit{Rand}       & \textit{TMCD}          & 0.36 \\
GeoQuery & Spider   & \textit{TMCD}          & \textit{TMCD}          & 0.68 & GeoQuery & SCAN     & \textit{TMCD-Rstr}  & \textit{MCD2}          & 0.36 \\
SCAN     & SCAN     & \textit{TurnLeftRcvcv} & \textit{Simple}        & 0.68 & GeoQuery & Spider   & \textit{Length}     & \textit{Rand}          & 0.35 \\
GeoQuery & GeoQuery & \textit{TMCD}          & \textit{Std}           & 0.68 & GeoQuery & SCAN     & \textit{TMCD-Rcvcv} & \textit{MCD2}          & 0.35 \\
COGS     & Spider   & \textit{TMCD}          & \textit{Std}           & 0.67 & Spider   & Spider   & \textit{Rand}       & \textit{Template}      & 0.35 \\
COGS     & GeoQuery & \textit{Std-Rstr}      & \textit{Length}        & 0.67 & GeoQuery & SCAN     & \textit{Length}     & \textit{Template}      & 0.35 \\
COGS     & GeoQuery & \textit{Template}      & \textit{Length}        & 0.67 & GeoQuery & SCAN     & \textit{Std}        & \textit{Jump}          & 0.35 \\
GeoQuery & GeoQuery & \textit{TMCD-Rcvcv}    & \textit{Template}      & 0.66 & GeoQuery & Spider   & \textit{Std}        & \textit{Rand}          & 0.35 \\
GeoQuery & GeoQuery & \textit{TMCD}          & \textit{Template}      & 0.65 & SCAN     & SCAN     & \textit{MCD2}       & \textit{Jump}          & 0.35 \\
GeoQuery & GeoQuery & \textit{TMCD-Rstr}     & \textit{TMCD}          & 0.65 & COGS     & GeoQuery & \textit{RandStr}    & \textit{TMCD}          & 0.34 \\
GeoQuery & GeoQuery & \textit{TMCD-Rstr}     & \textit{Std-Rstr}      & 0.65 & Spider   & Spider   & \textit{Length}     & \textit{TMCD}          & 0.34 \\
SCAN     & SCAN     & \textit{MCD2}          & \textit{Length}        & 0.64 & Spider   & Spider   & \textit{Length}     & \textit{Template}      & 0.34 \\
SCAN     & SCAN     & \textit{TurnLeftRStr}  & \textit{TurnLeftRcvcv} & 0.64 & COGS     & GeoQuery & \textit{Randcvcv}   & \textit{Std}           & 0.34 \\
COGS     & GeoQuery & \textit{Std}           & \textit{Std}           & 0.64 & SCAN     & Spider   & \textit{TurnLeft}   & \textit{Template}      & 0.34 \\
GeoQuery & Spider   & \textit{TMCD}          & \textit{Length}        & 0.63 & COGS     & GeoQuery & \textit{Randcvcv}   & \textit{Length}        & 0.34 \\
GeoQuery & GeoQuery & \textit{TMCD}          & \textit{Length}        & 0.63 & GeoQuery & SCAN     & \textit{TMCD}       & \textit{Jump}          & 0.33 \\
GeoQuery & GeoQuery & \textit{TMCD-Rcvcv}    & \textit{TMCD}          & 0.63 & COGS     & SCAN     & \textit{Std}        & \textit{MCD2}          & 0.33 \\
COGS     & GeoQuery & \textit{TMCD}          & \textit{Length}        & 0.63 & COGS     & GeoQuery & \textit{Randcvcv}   & \textit{Template}      & 0.33 \\
GeoQuery & Spider   & \textit{Template}      & \textit{Length}        & 0.63 & COGS     & GeoQuery & \textit{RandStr}    & \textit{Length}        & 0.32 \\
GeoQuery & GeoQuery & \textit{Length}        & \textit{Std}           & 0.62 & SCAN     & Spider   & \textit{TurnLeft}   & \textit{TMCD}          & 0.32 \\
GeoQuery & GeoQuery & \textit{Template}      & \textit{Length}        & 0.62 & GeoQuery & Spider   & \textit{Std}        & \textit{Length}        & 0.32 \\
GeoQuery & GeoQuery & \textit{Template}      & \textit{Std-Rcvcv}     & 0.62 & SCAN     & Spider   & \textit{Template}   & \textit{TMCD}          & 0.32 \\
GeoQuery & Spider   & \textit{Template}      & \textit{Std-Rstr}      & 0.6  & SCAN     & Spider   & \textit{MCD1}       & \textit{Length}        & 0.31 \\
COGS     & COGS     & \textit{RandStr}       & \textit{Std}           & 0.6  & GeoQuery & Spider   & \textit{Template}   & \textit{Rand}          & 0.31 \\
GeoQuery & GeoQuery & \textit{TMCD}          & \textit{Std-Rcvcv}     & 0.6  & GeoQuery & SCAN     & \textit{Template}   & \textit{Jump}          & 0.31 \\
COGS     & COGS     & \textit{Randcvcv}      & \textit{Std}           & 0.59 & GeoQuery & Spider   & \textit{TMCD}       & \textit{Rand}          & 0.31 \\
GeoQuery & Spider   & \textit{TMCD}          & \textit{Std-Rstr}      & 0.58 & SCAN     & Spider   & \textit{Template}   & \textit{Template}      & 0.31 \\
GeoQuery & GeoQuery & \textit{TMCD-Rcvcv}    & \textit{Length}        & 0.57 & GeoQuery & SCAN     & \textit{Std}        & \textit{Template}      & 0.3  \\
COGS     & GeoQuery & \textit{TMCD-Rcvcv}    & \textit{RandStr}       & 0.57 & GeoQuery & Spider   & \textit{Std-Rstr}   & \textit{Rand}          & 0.3  \\
SCAN     & SCAN     & \textit{MCD3}          & \textit{MCD2}          & 0.57 & COGS     & SCAN     & \textit{Randcvcv}   & \textit{TurnLeftRStr}  & 0.29 \\
COGS     & GeoQuery & \textit{Length}        & \textit{Std}           & 0.56 & SCAN     & SCAN     & \textit{TurnLeft}   & \textit{TurnLeftRcvcv} & 0.29 \\
COGS     & GeoQuery & \textit{TMCD}          & \textit{Std}           & 0.56 & GeoQuery & SCAN     & \textit{Template}   & \textit{Template}      & 0.28 \\
COGS     & GeoQuery & \textit{Template}      & \textit{Std}           & 0.56 & SCAN     & SCAN     & \textit{MCD2}       & \textit{TurnLeft}      & 0.28 \\
COGS     & GeoQuery & \textit{Std-Rcvcv}     & \textit{RandStr}       & 0.56 & COGS     & GeoQuery & \textit{Randcvcv}   & \textit{TMCD}          & 0.28 \\
GeoQuery & GeoQuery & \textit{TMCD-Rstr}     & \textit{Length}        & 0.55 & GeoQuery & SCAN     & \textit{Std}        & \textit{TurnLeft}      & 0.28 \\
COGS     & COGS     & \textit{Length}        & \textit{RandStr}       & 0.55 & COGS     & SCAN     & \textit{Length}     & \textit{MCD2}          & 0.28 \\
GeoQuery & SCAN     & \textit{Jump}          & \textit{Std-Rstr}      & 0.54 & GeoQuery & SCAN     & \textit{Length}     & \textit{TurnLeft}      & 0.28 \\
COGS     & GeoQuery & \textit{Length}        & \textit{Length}        & 0.54 & GeoQuery & SCAN     & \textit{TMCD}       & \textit{Template}      & 0.28 \\
GeoQuery & GeoQuery & \textit{Std-Rstr}      & \textit{Std}           & 0.54 & GeoQuery & Spider   & \textit{Std-Rstr}   & \textit{Length}        & 0.27      \\
\bottomrule
\end{tabular}
\end{adjustbox}
\caption{Concurrence Values.}
\label{tab:app:Concurrence}
\end{table*}

\begin{table*}[ht]
\centering
\scriptsize
\begin{adjustbox}{max width=\textwidth}
\begin{tabular}{llllllllll}
\toprule
\textbf{Dataset A} &
  \textbf{Dataset B} &
  \textbf{Split A} &
  \textbf{Split B} &
  \textbf{Concur} &
  \textbf{Dataset A} &
  \textbf{Dataset B} &
  \textbf{Split A} &
  \textbf{Split B} &
  \textbf{Concur} \\ \midrule
COGS     & Spider & \textit{Length}     & \textit{Rand}          & 0.27 & SCAN     & SCAN   & \textit{Jump}          & \textit{TurnLeftRcvcv} & 0.02  \\
GeoQuery & SCAN   & \textit{Std-Rstr}   & \textit{Template}      & 0.27 & COGS     & SCAN   & \textit{Std}           & \textit{MCD1}          & 0.02  \\
GeoQuery & SCAN   & \textit{Std-Rstr}   & \textit{MCD2}          & 0.27 & SCAN     & Spider & \textit{MCD3}          & \textit{Length}        & 0.02  \\
COGS     & SCAN   & \textit{RandStr}    & \textit{TurnLeftRStr}  & 0.27 & COGS     & SCAN   & \textit{Length}        & \textit{MCD3}          & 0.02  \\
COGS     & Spider & \textit{Length}     & \textit{Length}        & 0.26 & GeoQuery & SCAN   & \textit{TMCD-Rcvcv}    & \textit{TurnLeftRStr}  & 0.02  \\
SCAN     & SCAN   & \textit{Length}     & \textit{TurnLeft}      & 0.25 & SCAN     & SCAN   & \textit{MCD1}          & \textit{TurnLeft}      & 0.02  \\
GeoQuery & SCAN   & \textit{TMCD}       & \textit{TurnLeft}      & 0.24 & SCAN     & Spider & \textit{Length}        & \textit{Length}        & 0.01  \\
GeoQuery & Spider & \textit{Template}   & \textit{Length}        & 0.24 & COGS     & SCAN   & \textit{Length}        & \textit{Length}        & 0.01  \\
COGS     & SCAN   & \textit{Randcvcv}   & \textit{Simple}        & 0.24 & SCAN     & SCAN   & \textit{Simple}        & \textit{MCD3}          & 0.01  \\
SCAN     & SCAN   & \textit{MCD1}       & \textit{Template}      & 0.24 & SCAN     & Spider & \textit{Simple}        & \textit{Length}        & 0.01  \\
SCAN     & SCAN   & \textit{TurnLeft}   & \textit{TurnLeftRStr}  & 0.23 & SCAN     & SCAN   & \textit{TurnLeft}      & \textit{Template}      & 0.01  \\
GeoQuery & SCAN   & \textit{Template}   & \textit{TurnLeft}      & 0.23 & SCAN     & SCAN   & \textit{Simple}        & \textit{Jump}          & 0.0   \\
GeoQuery & Spider & \textit{TMCD}       & \textit{Length}        & 0.23 & SCAN     & Spider & \textit{TurnLeft}      & \textit{Rand}          & -0.0  \\
GeoQuery & Spider & \textit{Length}     & \textit{Length}        & 0.23 & COGS     & SCAN   & \textit{Randcvcv}      & \textit{Length}        & -0.01 \\
GeoQuery &
  Spider &
  \textit{TMCD-Rcvcv} &
  \textit{Length} &
  0.22 &
  GeoQuery &
  SCAN &
  \textit{Std-Rcvcv} &
  \textit{TurnLeftRStr} &
  -0.01 \\
SCAN     & SCAN   & \textit{Length}     & \textit{Jump}          & 0.22 & GeoQuery & SCAN   & \textit{Std}           & \textit{MCD1}          & -0.02 \\
COGS     & SCAN   & \textit{Length}     & \textit{Template}      & 0.22 & SCAN     & Spider & \textit{MCD1}          & \textit{Template}      & -0.02 \\
GeoQuery & Spider & \textit{TMCD-Rstr}  & \textit{Length}        & 0.22 & GeoQuery & SCAN   & \textit{TMCD}          & \textit{MCD1}          & -0.02 \\
GeoQuery & Spider & \textit{TMCD-Rcvcv} & \textit{Rand}          & 0.22 & COGS     & SCAN   & \textit{Std}           & \textit{TurnLeftRcvcv} & -0.02 \\
GeoQuery & Spider & \textit{TMCD-Rstr}  & \textit{Rand}          & 0.21 & COGS     & SCAN   & \textit{RandStr}       & \textit{MCD1}          & -0.02 \\
COGS     & SCAN   & \textit{RandStr}    & \textit{Simple}        & 0.21 & COGS     & SCAN   & \textit{Std}           & \textit{Simple}        & -0.03 \\
SCAN     & SCAN   & \textit{MCD1}       & \textit{Jump}          & 0.21 & COGS     & SCAN   & \textit{Length}        & \textit{TurnLeftRStr}  & -0.03 \\
SCAN     & Spider & \textit{MCD2}       & \textit{Template}      & 0.2  & SCAN     & Spider & \textit{TurnLeftRStr}  & \textit{Length}        & -0.03 \\
GeoQuery & SCAN   & \textit{Std}        & \textit{MCD2}          & 0.2  & GeoQuery & SCAN   & \textit{TMCD}          & \textit{MCD3}          & -0.03 \\
COGS     & SCAN   & \textit{RandStr}    & \textit{TurnLeftRcvcv} & 0.2  & SCAN     & Spider & \textit{MCD1}          & \textit{TMCD}          & -0.03 \\
SCAN     & SCAN   & \textit{Simple}     & \textit{TurnLeft}      & 0.2  & GeoQuery & SCAN   & \textit{TMCD-Rcvcv}    & \textit{Simple}        & -0.04 \\
SCAN     & Spider & \textit{MCD1}       & \textit{Rand}          & 0.19 & GeoQuery & SCAN   & \textit{Template}      & \textit{MCD3}          & -0.04 \\
SCAN     & Spider & \textit{MCD2}       & \textit{TMCD}          & 0.19 & GeoQuery & SCAN   & \textit{TMCD}          & \textit{Length}        & -0.04 \\
GeoQuery & Spider & \textit{Std-Rcvcv}  & \textit{Rand}          & 0.18 & COGS     & SCAN   & \textit{RandStr}       & \textit{Length}        & -0.04 \\
GeoQuery & SCAN   & \textit{TMCD-Rcvcv} & \textit{Template}      & 0.18 & SCAN     & SCAN   & \textit{MCD3}          & \textit{Template}      & -0.05 \\
COGS     & Spider & \textit{RandStr}    & \textit{Rand}          & 0.18 & GeoQuery & SCAN   & \textit{TMCD-Rcvcv}    & \textit{TurnLeftRcvcv} & -0.05 \\
GeoQuery & SCAN   & \textit{TMCD-Rstr}  & \textit{Template}      & 0.18 & GeoQuery & SCAN   & \textit{Std-Rcvcv}     & \textit{Simple}        & -0.06 \\
SCAN     & SCAN   & \textit{MCD3}       & \textit{Jump}          & 0.18 & GeoQuery & SCAN   & \textit{Std}           & \textit{MCD3}          & -0.06 \\
GeoQuery & SCAN   & \textit{TMCD}       & \textit{MCD2}          & 0.18 & SCAN     & Spider & \textit{MCD3}          & \textit{Template}      & -0.06 \\
SCAN     & Spider & \textit{MCD2}       & \textit{Length}        & 0.18 & COGS     & SCAN   & \textit{Randcvcv}      & \textit{MCD3}          & -0.06 \\
GeoQuery & SCAN   & \textit{Template}   & \textit{MCD2}          & 0.17 & GeoQuery & SCAN   & \textit{TMCD-Rstr}     & \textit{TurnLeftRStr}  & -0.06 \\
SCAN     & SCAN   & \textit{MCD3}       & \textit{TurnLeft}      & 0.17 & GeoQuery & SCAN   & \textit{Template}      & \textit{Length}        & -0.06 \\
COGS     & Spider & \textit{Std}        & \textit{Rand}          & 0.17 & COGS     & SCAN   & \textit{Length}        & \textit{Simple}        & -0.07 \\
GeoQuery & Spider & \textit{Std-Rcvcv}  & \textit{Length}        & 0.17 & SCAN     & SCAN   & \textit{Length}        & \textit{Template}      & -0.07 \\
COGS     & Spider & \textit{RandStr}    & \textit{Length}        & 0.15 & SCAN     & Spider & \textit{MCD3}          & \textit{TMCD}          & -0.07 \\
GeoQuery & SCAN   & \textit{Std-Rstr}   & \textit{TurnLeftRStr}  & 0.15 & GeoQuery & SCAN   & \textit{Std}           & \textit{Length}        & -0.07 \\
COGS     & SCAN   & \textit{RandStr}    & \textit{MCD2}          & 0.15 & SCAN     & Spider & \textit{Length}        & \textit{Template}      & -0.07 \\
SCAN     & SCAN   & \textit{Length}     & \textit{TurnLeftRcvcv} & 0.14 & COGS     & SCAN   & \textit{RandStr}       & \textit{MCD3}          & -0.07 \\
COGS     & SCAN   & \textit{Std}        & \textit{Length}        & 0.14 & GeoQuery & SCAN   & \textit{Length}        & \textit{MCD3}          & -0.08 \\
COGS     & SCAN   & \textit{RandStr}    & \textit{Template}      & 0.14 & SCAN     & Spider & \textit{MCD3}          & \textit{Rand}          & -0.08 \\
COGS     & Spider & \textit{Std}        & \textit{Length}        & 0.14 & COGS     & SCAN   & \textit{Randcvcv}      & \textit{MCD1}          & -0.09 \\
GeoQuery & SCAN   & \textit{TMCD-Rcvcv} & \textit{Length}        & 0.14 & GeoQuery & SCAN   & \textit{Length}        & \textit{Length}        & -0.09 \\
COGS     & SCAN   & \textit{Std}        & \textit{Template}      & 0.13 & SCAN     & Spider & \textit{Length}        & \textit{TMCD}          & -0.09 \\
GeoQuery & SCAN   & \textit{Std-Rcvcv}  & \textit{Template}      & 0.13 & SCAN     & SCAN   & \textit{MCD1}          & \textit{TurnLeftRStr}  & -0.09 \\
COGS     & SCAN   & \textit{Randcvcv}   & \textit{MCD2}          & 0.13 & SCAN     & Spider & \textit{Length}        & \textit{Rand}          & -0.1  \\
GeoQuery & SCAN   & \textit{TMCD-Rstr}  & \textit{Length}        & 0.12 & SCAN     & Spider & \textit{TurnLeftRStr}  & \textit{Template}      & -0.11 \\
SCAN &
  SCAN &
  \textit{Length} &
  \textit{TurnLeftRStr} &
  0.12 &
  GeoQuery &
  SCAN &
  \textit{Std-Rcvcv} &
  \textit{TurnLeftRcvcv} &
  -0.11 \\
COGS     & SCAN   & \textit{Std}        & \textit{MCD3}          & 0.12 & SCAN     & SCAN   & \textit{Simple}        & \textit{MCD1}          & -0.11 \\
GeoQuery & SCAN   & \textit{Std-Rcvcv}  & \textit{Length}        & 0.12 & SCAN     & Spider & \textit{TurnLeftRStr}  & \textit{TMCD}          & -0.12 \\
GeoQuery & SCAN   & \textit{TMCD-Rstr}  & \textit{MCD1}          & 0.11 & SCAN     & Spider & \textit{Simple}        & \textit{Rand}          & -0.12 \\
COGS     & Spider & \textit{Randcvcv}   & \textit{Rand}          & 0.11 & COGS     & SCAN   & \textit{Length}        & \textit{TurnLeftRcvcv} & -0.12 \\
GeoQuery & SCAN   & \textit{TMCD-Rcvcv} & \textit{MCD3}          & 0.11 & GeoQuery & SCAN   & \textit{TMCD-Rstr}     & \textit{Simple}        & -0.13 \\
GeoQuery & SCAN   & \textit{Std-Rcvcv}  & \textit{MCD3}          & 0.11 & SCAN     & SCAN   & \textit{MCD1}          & \textit{TurnLeftRcvcv} & -0.13 \\
GeoQuery & SCAN   & \textit{TMCD-Rstr}  & \textit{MCD3}          & 0.11 & SCAN     & SCAN   & \textit{Simple}        & \textit{Template}      & -0.13 \\
GeoQuery & SCAN   & \textit{Std-Rcvcv}  & \textit{MCD1}          & 0.1  & SCAN     & Spider & \textit{TurnLeftRStr}  & \textit{Rand}          & -0.14 \\
GeoQuery & SCAN   & \textit{Std-Rstr}   & \textit{MCD1}          & 0.1  & GeoQuery & SCAN   & \textit{TMCD-Rstr}     & \textit{TurnLeftRcvcv} & -0.14 \\
GeoQuery & SCAN   & \textit{Std-Rstr}   & \textit{Simple}        & 0.09 & SCAN     & Spider & \textit{Simple}        & \textit{Template}      & -0.15 \\
GeoQuery & SCAN   & \textit{Std-Rstr}   & \textit{Length}        & 0.09 & SCAN     & SCAN   & \textit{TurnLeftRStr}  & \textit{Template}      & -0.15 \\
GeoQuery &
  SCAN &
  \textit{Std-Rstr} &
  \textit{TurnLeftRcvcv} &
  0.08 &
  SCAN &
  Spider &
  \textit{TurnLeftRcvcv} &
  \textit{Length} &
  -0.15 \\ 
COGS     & SCAN   & \textit{Randcvcv}   & \textit{Template}      & 0.08 & GeoQuery & SCAN   & \textit{Std}           & \textit{TurnLeftRStr}  & -0.15 \\
SCAN     & SCAN   & \textit{MCD2}       & \textit{TurnLeftRStr}  & 0.08 & SCAN     & Spider & \textit{Simple}        & \textit{TMCD}          & -0.16 \\
SCAN     & SCAN   & \textit{Simple}     & \textit{Length}        & 0.08 & GeoQuery & SCAN   & \textit{Length}        & \textit{MCD1}          & -0.18 \\
COGS     & Spider & \textit{Randcvcv}   & \textit{Length}        & 0.07 & GeoQuery & SCAN   & \textit{Std}           & \textit{Simple}        & -0.19 \\
SCAN     & SCAN   & \textit{MCD2}       & \textit{TurnLeftRcvcv} & 0.07 & SCAN     & Spider & \textit{TurnLeftRcvcv} & \textit{Template}      & -0.2  \\
SCAN     & SCAN   & \textit{MCD3}       & \textit{TurnLeftRcvcv} & 0.06 & GeoQuery & SCAN   & \textit{TMCD}          & \textit{TurnLeftRStr}  & -0.21 \\
GeoQuery & SCAN   & \textit{Length}     & \textit{MCD2}          & 0.06 & GeoQuery & SCAN   & \textit{Template}      & \textit{TurnLeftRStr}  & -0.21 \\
SCAN     & SCAN   & \textit{Simple}     & \textit{MCD2}          & 0.05 & SCAN     & Spider & \textit{TurnLeftRcvcv} & \textit{TMCD}          & -0.22 \\
GeoQuery & SCAN   & \textit{TMCD-Rcvcv} & \textit{MCD1}          & 0.05 & GeoQuery & SCAN   & \textit{Length}        & \textit{TurnLeftRStr}  & -0.24 \\
SCAN     & SCAN   & \textit{MCD3}       & \textit{TurnLeftRStr}  & 0.05 & GeoQuery & SCAN   & \textit{Std}           & \textit{TurnLeftRcvcv} & -0.25 \\
SCAN     & SCAN   & \textit{Jump}       & \textit{TurnLeftRStr}  & 0.05 & SCAN     & SCAN   & \textit{TurnLeftRcvcv} & \textit{Template}      & -0.26 \\
SCAN     & Spider & \textit{MCD2}       & \textit{Rand}          & 0.05 & GeoQuery & SCAN   & \textit{TMCD}          & \textit{Simple}        & -0.26 \\
SCAN     & Spider & \textit{TurnLeft}   & \textit{Length}        & 0.05 & GeoQuery & SCAN   & \textit{Template}      & \textit{Simple}        & -0.27 \\
GeoQuery & SCAN   & \textit{Std-Rstr}   & \textit{MCD3}          & 0.05 & SCAN     & Spider & \textit{TurnLeftRcvcv} & \textit{Rand}          & -0.27 \\
SCAN     & SCAN   & \textit{MCD2}       & \textit{Template}      & 0.04 & GeoQuery & SCAN   & \textit{Length}        & \textit{TurnLeftRcvcv} & -0.28 \\
COGS     & SCAN   & \textit{Length}     & \textit{MCD1}          & 0.04 & GeoQuery & SCAN   & \textit{TMCD}          & \textit{TurnLeftRcvcv} & -0.29 \\
COGS     & SCAN   & \textit{Std}        & \textit{TurnLeftRStr}  & 0.03 & GeoQuery & SCAN   & \textit{Template}      & \textit{TurnLeftRcvcv} & -0.3  \\
GeoQuery & SCAN   & \textit{Template}   & \textit{MCD1}          & 0.02 & GeoQuery & SCAN   & \textit{Length}        & \textit{Simple}        & -0.3     
\\
\bottomrule
\end{tabular}
\end{adjustbox}
\caption{Concurrence Values (Cont).}
\label{tab:app:allconur}
\end{table*}

\section{Mistakes that model make in both random splits and generalization splits}
\label{app:sec:mistakes}
\begin{table*}[ht]
\scriptsize
\centering
\begin{adjustbox}{max width=\textwidth}
\begin{tabular}{p{7em}p{0.8\textwidth}}
\toprule
\textbf{Example 1.}  & BART on GeoQuery \textit{standard} and \textit{template} \\
\textbf{Input:} & what are the highest points of all the states \\
\textbf{Output:} & answer ( highest ( intersection ( place , loc\_2 ( state ) ) ) ) \\
\textbf{Prediction:} & answer ( highest ( intersection ( place , loc\_2 ( state ) ) ) ) \hlc[cyan!50]{)}  \\
\hline
\textbf{Input:} & what is the adjacent state of m0 \\
\textbf{Output:} & answer ( intersection ( state , next\_to\_2 ( m0 ) ) ) \\
\textbf{Prediction:} & answer ( intersection ( state , next\_to\_2 ( m0 ) ) ) \hlc[cyan!50]{)} \\ \midrule
\textbf{Example 2.}  & BTG on GeoQuery \textit{simple} and \textit{TurnLeft} \\
\textbf{Input:} & run left thrice and look opposite right thrice \\
\textbf{Output:} & TURN\_LEFT RUN TURN\_LEFT RUN TURN\_LEFT RUN TURN\_\hlc[cyan!50]{RIGHT} TURN\_\hlc[cyan!50]{RIGHT} LOOK TURN\_\hlc[cyan!50]{RIGHT} TURN\_\hl{RIGHT} LOOK TURN\_\hlc[cyan!50]{RIGHT} TURN\_\hlc[cyan!50]{RIGHT} I\_LOOK \\
\textbf{Prediction:} & TURN\_LEFT RUN TURN\_LEFT RUN TURN\_LEFT RUN TURN\_\hlc[cyan!50]{LEFT} TURN\_\hlc[cyan!50]{LEFT} LOOK TURN\_\hlc[cyan!50]{LEFT} TURN\_\hlc[cyan!50]{LEFT} LOOK TURN\_\hlc[cyan!50]{LEFT} TURN\_\hlc[cyan!50]{LEFT} LOOK \\ \hline
\textbf{Input:} & look right after turn left \\
\textbf{Output:} & TURN\_LEFT TURN\_\hlc[cyan!50]{RIGHT} LOOK \\
\textbf{Prediction:} & TURN\_LEFT TURN\_\hlc[cyan!50]{LEFT} LOOK \\
\bottomrule
\end{tabular}
\end{adjustbox}
\caption{Examples of instance where the model makes both mistakes in random split and generalization split. The first instance is the output of BART on \textit{standard} split of GeoQuery, and the second entry is BART making a similar mistake on \textit{template} split of GeoQuery; the second instance is output of BTG on \textit{simple} split of SCAN, and a similar instance making the same directional mistake on the \textit{TurnLeft} split.}
\label{tab:harmless-example}

\end{table*}
The in-distribution performance may also be a confounder when at least one of the models does not perform as well on an in-distribution test set, or in a random split of the data.
Qualitatively, we observe that models sometimes make the same trivial mistakes in both a random split and a generalization split, making the resulting raw metric unrepresentative of compositionality.
For example, BART makes mistakes on parentheses, adding or dropping them on both standard split and generalization splits of GeoQuery (Table~\ref{tab:harmless-example}); BTG cannot tell left from right in the \textit{simple} split of SCAN, and the same type of mistake continues to appear in the \textit{template} split.
While simple mistakes like these and the space tokenization issue mentioned in Section~\ref{sec:expsetup:eval} can be easily resolved by adopting a post-processing protocol or rules to ignore when computing EM, other types of less identifiable errors may also be present and harder to patch.
Since many of the models do not achieve near-perfect performance on the random splits, to what extent they make the mistakes in the standard split again in the generalization splits requires further research.

We also include a Genbench evaluation card \cite{Hupkes2023} in \cref{tab:eval_card}.
\newcommand{\tabularwidth}{\textwidth}

\newcommand{\expone}{$\square$}
\newcommand{\exptwo}{$\bigtriangleup$}
\newcommand{\expthree}{$\bigcirc$}
\newcommand{\expfour}{$\odot$}
        
\renewcommand{\arraystretch}{1.1}         
\setlength{\tabcolsep}{0mm}        
\begin{table*}[ht]
\begin{adjustbox}{max width=\textwidth}

\begin{tabular}{|p{\tabularwidth}<{\centering}|}         
\hline
               
\rowcolor{gray!60}               
\textbf{Motivation} \\               
\footnotesize
\begin{tabular}{p{0.25\tabularwidth}<{\centering} p{0.25\tabularwidth}<{\centering} p{0.25\tabularwidth}<{\centering} p{0.25\tabularwidth}<{\centering}}                        
\textit{Practical} & \textit{Cognitive} & \textit{Intrinsic} & \textit{Fairness}\\
& \expone\hspace{0.8mm}\exptwo\hspace{0.8mm}\expthree\hspace{0.8mm}\expfour\hspace{0.8mm}		%
& 		%
& 		%

\vspace{2mm} \\
\end{tabular}\\
               
\rowcolor{gray!60}               
\textbf{Generalisation type} \\               
\footnotesize
\begin{tabular}{m{0.21\tabularwidth}<{\centering} m{0.2\tabularwidth}<{\centering} m{0.13\tabularwidth}<{\centering} m{0.13\tabularwidth}<{\centering} m{0.13\tabularwidth}<{\centering} m{0.2\tabularwidth}<{\centering}}                   
\textit{Compositional} & \textit{Structural} & \textit{Cross Task} & \textit{Cross Language} & \textit{Cross Domain} & \textit{Robustness}\\
\expone\hspace{0.8mm}\exptwo\hspace{0.8mm}\expthree\hspace{0.8mm}\expfour\hspace{0.8mm}		%
& 		%
& 		%
& 		%
& 		%
& 		%

\vspace{2mm} \\
\end{tabular}\\
             
\rowcolor{gray!60}             
\textbf{Shift type} \\             
\footnotesize
\begin{tabular}{p{0.25\tabularwidth}<{\centering} p{0.25\tabularwidth}<{\centering} p{0.25\tabularwidth}<{\centering} p{0.25\tabularwidth}<{\centering}}                        
\textit{Covariate} & \textit{Label} & \textit{Full} & \textit{Assumed}\\  
\expone\hspace{0.8mm}\exptwo\hspace{0.8mm}\expthree\hspace{0.8mm}\expfour\hspace{0.8mm}		%
& 		%
& 		%
& 		%

\vspace{2mm} \\
\end{tabular}\\
             
\rowcolor{gray!60}             
\textbf{Shift source} \\             
\footnotesize
\begin{tabular}{p{0.25\tabularwidth}<{\centering} p{0.25\tabularwidth}<{\centering} p{0.25\tabularwidth}<{\centering} p{0.25\tabularwidth}<{\centering}}                          
\textit{Naturally occuring} & \textit{Partitioned natural} & \textit{Generated shift} & \textit{Fully generated}\\
& \expone\hspace{0.8mm}\exptwo\hspace{0.8mm}		%
& 		%
& \hspace{11mm}\expthree\hspace{0.8mm}\expfour\hspace{0.8mm}		%

\vspace{2mm} \\
\end{tabular}\\
             
\rowcolor{gray!60}             
\textbf{Shift locus}\\             
\footnotesize
\begin{tabular}{p{0.25\tabularwidth}<{\centering} p{0.25\tabularwidth}<{\centering} p{0.25\tabularwidth}<{\centering} p{0.25\tabularwidth}<{\centering}}                         
\textit{Train--test} & \textit{Finetune train--test} & \textit{Pretrain--train} & \textit{Pretrain--test}\\
\expone\hspace{0.8mm}\expthree\hspace{0.8mm}		%
& \exptwo\hspace{0.8mm}\expfour\hspace{0.8mm}		%
& 		%
& 		%

\vspace{2mm} \\
\end{tabular} \\
\hline

\end{tabular}
\end{adjustbox}
\caption{A GenBench evaluation card \citep{Hupkes2023} that summarizes our experiments. \expone = Experiments of LSTM and Transformer on GeoQuery and Spider; \exptwo = Experiments of T5 and BART on GeoQuery and Spider; \expthree = Experiments of LSTM and Transformer on COGS and SCAN; \expfour = Experiments of T5 and BART on COGS and SCAN.}
\label{tab:eval_card}
\end{table*}

\section{Limitations}

While we explore the consequences of the modeling approach on concurrence, we have focused mainly on models trained from scratch to perform compositional generalization or pretrained models which have been finetuned. Another possible area of investigation would be to explore the extent to which a model's compositional generalization abilities also transfer to in-context evaluations \citep{hosseini-etal-2022-compositional}. We leave this question for future work.

\end{document}